\documentclass[10pt,twocolumn,letterpaper]{article}

\usepackage{cvpr}              
\usepackage{graphicx}
\usepackage{amsmath}
\usepackage{amssymb}
\usepackage{booktabs}
\usepackage{multirow}
\usepackage{float}
\usepackage{amsthm} \usepackage{color}
\usepackage{xspace}
\usepackage{bbm}
\usepackage{array}
\usepackage{caption}
\usepackage{subcaption}
\usepackage{colortbl}
\usepackage{nicefrac}
\usepackage{url}
\usepackage{tikz}
\usepackage{comment}
\usepackage[accsupp]{axessibility}

\usepackage[pagebackref,breaklinks,colorlinks]{hyperref}

\usepackage[capitalize]{cleveref}
\crefname{section}{Sec.}{Secs.}
\Crefname{section}{Section}{Sections}
\Crefname{table}{Table}{Tables}
\crefname{table}{Tab.}{Tabs.}
\crefname{table}{Tab.}{Tabs.}

\newcommand{\vqatwo}{VQA v2\xspace}
\newcommand{\advqa}{AdVQA\xspace}
\newcommand{\ours}{LYP\xspace}
\newcommand{\fullours}{Learning from Your Peers\xspace}

\newcommand{\myparagraph}[1]{\noindent\textbf{#1}}

\definecolor{light-gray}{gray}{0.95} 
\definecolor{grey}{gray}{0.8} 
\newcommand\grey[1]{{\color{grey}{#1}}}

\newcommand{\symcovatr}{$\mathcal{C}@\mathcal{R}$\xspace}
\newcommand{\covatr}[1]{$\mathcal{C}@$#1\%\xspace}
\newcommand{\symeffrel}{$\Phi_c$\xspace}
\newcommand{\effrel}[1]{$\Phi_{#1}$\xspace}
\newcommand{\vqa}{VQA v2\xspace}
\newcommand{\train}{A\xspace}
\newcommand{\dev}{B\xspace}
\newcommand{\traindev}{A+B\xspace}

\newcommand{\val}{Val\xspace}
\newcommand{\test}{Test\xspace}
\newcommand{\clipvil}{CLIP-ViL\xspace}
\newcommand{\ofa}{OFA\xspace}
\newcommand{\ofabase}{OFA-Base\xspace}
\newcommand{\ofalarge}{OFA-Large\xspace}
\newcommand{\maxprob}{MaxProb\xspace}
\newcommand{\selector}{Selector\xspace}

\newcommand{\authorskip}{\hspace{3mm}}

\def\cvprPaperID{6665} \def\confName{CVPR}
\def\confYear{2023}

\begin{document}

\title{
Improving Selective Visual Question Answering by Learning from Your Peers
 }

\author{
Corentin Dancette$^{1,3
\dagger*}$ \authorskip Spencer Whitehead$^{1*}$ \authorskip Rishabh Maheshwary$^{1}$ \authorskip Ramakrishna Vedantam$^{1}$ \\
Stefan Scherer$^{2}$ \authorskip Xinlei Chen$^{1}$ \authorskip Matthieu Cord$^{3,4}$ \authorskip Marcus Rohrbach$^{1}$ \\[2mm]
$^{1}$FAIR, Meta AI \authorskip\authorskip $^{2}$Reality Labs Research, Meta \authorskip\authorskip $^{3}$Sorbonne Universit\'e \authorskip\authorskip$^4$Valeo.ai
}
\maketitle

\newcommand\blfootnote[1]{  \begingroup
  \renewcommand\thefootnote{}\footnote{#1}  \addtocounter{footnote}{-1}  \endgroup
}

\renewcommand{\thefootnote}{\fnsymbol{footnote}}
\footnotetext[1]{Equal contribution.}
\footnotetext[2]{Work primarily done during internship at FAIR.}
\blfootnote{Code: {\tiny \url{https://github.com/facebookresearch/selective-vqa_ood}}}
\renewcommand{\thefootnote}{\arabic{footnote}}
\setcounter{footnote}{0}

\begin{abstract}

Despite advances in Visual Question Answering (VQA), the ability of models to assess their own correctness remains under-explored. Recent work has shown that VQA models, out-of-the-box, can have difficulties abstaining from answering when they are wrong.
The option to abstain, also called Selective Prediction, is highly relevant when deploying systems to users who must trust the system's output (e.g., VQA assistants for users with visual impairments).
For such scenarios, abstention can be especially important as users may provide out-of-distribution (OOD) or adversarial inputs that make incorrect answers more likely.
In this work, we explore Selective VQA in both in-distribution (ID) and OOD scenarios, where models are presented with mixtures of ID and OOD data.
The goal is to maximize the number of questions answered while minimizing the risk of error on those questions.
We propose a simple yet effective \emph{\fullours (\ours)} approach for training multimodal selection functions for making abstention decisions.
Our approach uses predictions from models trained on distinct subsets of the training data as targets for optimizing a Selective VQA model.
It does not require additional manual labels or held-out data and provides a signal for identifying examples that are easy/difficult to generalize to.
In our extensive evaluations, we show this benefits a number of models across different architectures and scales.
Overall, for ID, we reach 32.92\% in the selective prediction metric coverage at 1\% risk of error (\covatr{1}) which doubles the previous best coverage of 15.79\% on this task.
For mixed ID/OOD, using models' softmax confidences for abstention decisions performs very poorly, answering $<$5\% of questions at 1\% risk of error even when faced with only 10\% OOD examples, but a learned selection function with \ours can increase that to 25.38\% \covatr{1}.

\end{abstract}

\section{Introduction}\label{sec:intro}

Recent successes of deep learning models for multimodal tasks have created the potential for many exciting real-world applications that require a large degree of reliability, such as providing assistance to users with visual impairments~\cite{gurari2018vizwiz,sidorov2020textcaps}.
However, with these novel, high-stakes applications come responsibilities towards the users, as well as the need to revise problem setups and the general approach to evaluating model performance. One particularly important consideration when developing models for real-world applications is \emph{reliability}, i.e., the ability of the model to avoid making errors when facing uncertainty.

\begin{figure}[t]
    \centering
        \includegraphics[width=\linewidth]{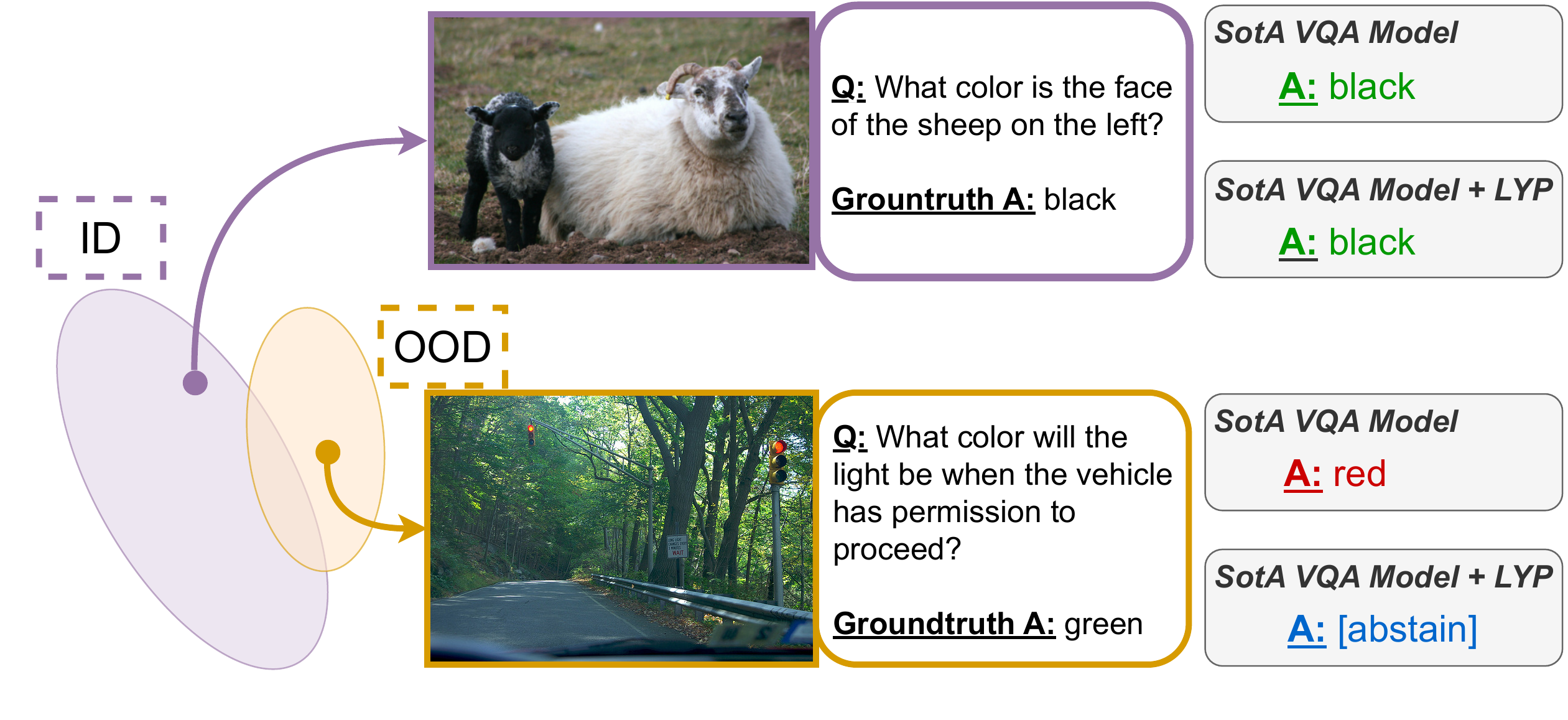}
    \caption{VQA Models are able to answer straightforward ID questions, as in the top example where a SotA model ~\cite{wang2022ofa} with and without our \fullours (\ours) approach answers correctly. However, difficult OOD examples can arise, like the bottom example. With \ours, the model is able to abstain from answering to avoid outputting the incorrect answer, whereas the existing model is overconfident and outputs the answer anyways.}
        \label{fig:teaser}
\end{figure}

One way to approach reliability is to frame the problem as a selective prediction task~\cite{chow1957optimum,elyaniv2010foundations,whitehead2022reliable}.
In selective prediction, models are able to either output an answer or abstain from answering (i.e., effectively saying ``\textit{I don't know}'') based on the model's confidence/uncertainty in order to avoid making incorrect predictions.
A prevalent cause of such incorrect predictions in real-world settings is distribution shifts~\cite{degrave2021aicovid,maartensson2020clinicalreliability,geirhos2020shortcut}, where the test environment may differ from the training environment and models could encounter a wide variety of input examples at test time that may not satisfy the independent and identically distributed assumption often made by practitioners when developing models.
This is especially true in open-ended tasks like Visual Question Answering (VQA) where models may receive adversarial, out-of-distribution (OOD) inputs that are difficult to answer correctly.
For example, in \cref{fig:teaser}, a model is asked a question that requires background knowledge that it simply does not possess.
While the ability to answer open-ended questions has been a point of focus in VQA, having a model perfectly answer all questions, ID and OOD, is likely unattainable~\cite{geiger-etal-2019-posing,kamath-etal-2020-selective}.
Therefore, framing this problem as a selective prediction task provides an avenue to handle such OOD examples more gracefully as the model can abstain from answering on many of these inputs, while still attempting to answer as many questions as possible.
Doing this requires models to recognize OOD examples for abstention decisions (OOD detection) and generalize to OOD examples (OOD generalization) in order to make predictions on examples that the model will get right.

However, previous evaluations for selective prediction in VQA~\cite{whitehead2022reliable} have been done on ID data, where the questions and images all come from the \vqatwo dataset~\cite{goyal2017vqav2}.
In NLP, there are efforts on selective prediction with OOD examples~\cite{kamath-etal-2020-selective,varshney2022investigating}, although they tend to not address some practical considerations, such as assuming access to OOD data or threshold generalization.
More broadly, selective prediction and OOD generalization have largely been studied as independent problems in the literature~\cite{tran2022plex}.

In this work, we explore selective prediction for VQA with distribution shifts, where we present models with mixtures of both ID and OOD examples, and measure the ability of different approaches to optimize answering as many questions as possible while maintaining a low risk of error (or high accuracy) on those questions.
We perform extensive experiments on \vqatwo~\cite{goyal2017vqav2} as our ID data and \advqa~\cite{sheng2021advqa} as our adversarial, OOD data.

We evaluate a number of state-of-the-art approaches to this problem and find that existing models' softmax probabilities are generally poor confidence estimates for abstention decisions on OOD data, leading models to answer $<$5\% of questions to achieve 1\% risk of error in some settings.
Further, we show that training a selection function~\cite{whitehead2022reliable} improves performance ID and OOD, but integrating features from OOD detection methods as well as augmenting with known-OOD data (i.e., OOD data different from the unknown target distribution) do not improve beyond simply training this selection function on ID data.
However, we observe that existing methods for training multimodal selection functions can require a held-out dataset in order to be most effective.

Therefore, we propose a \fullours (\ours) approach that alleviates the need for held-out data while also allowing both the VQA model and selection function to learn from the additional data that would have been withheld.
\ours works by breaking the training data into $N$ subsets and training different VQA models on distinct combinations of $N-1$ subsets, leaving one subset out at a time.
Our approach then uses these trained models to predict answers on their respective $N^{\text{th}}$ left-out subsets.
We recombine this data into an updated training set that has predictions from the different models.
We utilize these predictions and the associated accuracies as labels to train a multimodal selection function, which learns to predict the accuracies.
By using predictions on the training data from models that have not seen these examples, our approach provides a signal for which examples in the training data can be generalized to for a given model class, and which are too hard and should be abstained on.

Overall, our contributions are: We present an evaluation benchmark for Selective VQA on both ID and OOD data.
We show that model and data scaling are important factors for selective prediction and evaluate multiple baselines from prior works.
Finally, we propose \ours and demonstrate that it can benefit performance over standard selection function training in both ID and mixed ID/OOD settings.

\section{Related Work}\label{sec:relatedwork}

\myparagraph{Visual Question Answering.} VQA is a popular multimodal task that requires an understanding of both vision and language modalities to predict answers.
There are many standard datasets~\cite{antol2015vqa,hudson2019gqa,goyal2017vqav2,gurari2018vizwiz} and models for this task~\cite{anderson2018butd,jiang2018pythia,jiang2020defense,li2019visualbert,lu2019vilbert,shen2021clipvil,tan2019lxmert,wang2022ofa}.
In our work, we employ recent state-of-the-art models~\cite{shen2021clipvil,wang2022ofa} as our backbones to explore selective VQA.

\myparagraph{OOD VQA.}
Multiple works have demonstrated that VQA models often rely on shortcuts and do not generalize well on OOD data.
VQA-CP~\cite{agrawal2018vqacp} shows that VQA models rely on superficial correlation and lack image grounding.
GQA-OOD~\cite{kervadec2021gqaood} introduces an OOD benchmark that increases question diversity by including questions from various groups.
VQA-CE~\cite{dancette2021beyond} takes a step further and considers biases on both questions and images.
AdVQA~\cite{sheng2021advqa} and A-VQA~\cite{li2021avqa} are recently introduced VQA benchmarks that comprise adversarial questions using human and model-in-the-loop procedures to generate adversarial examples.
Other datasets require different abilities, such as TextVQA~\cite{singh2019textvqa} which contains questions requiring reading text in the image, or OK-VQA~\cite{marino2019okvqa} which requires external knowledge.
Methods to overcome difficulties related to OOD data include~\cite{cadene2019rubi}, which tackles unimodal biases, and \cite{ramakrishnan2018overcoming}, which improves image grounding using adversarial regularization.
Recently, \cite{agrawal2022rethinking} performs cross-dataset evaluations where VQA models exhibit poor generalization.

\myparagraph{Selective prediction \& reliability.}
Recently, \cite{whitehead2022reliable} explore Selective Prediction for VQA with ID data.
They experiment with different selectors on top of the base VQA model for improving their reliability on the VQA task.
\cite{varshney2022investigating} investigates selective prediction approaches across several NLP tasks in ID, OOD, and adversarial settings.
Specifically, they trained a selector (MLP) on top of the base model on a held-out split and used the selector's confidence scores to either answer or abstain from answering and improved risk, and coverage metrics compared to MaxProb.
\cite{corbiere2019addressing} studies failure prediction in deep neural networks by training a confidence model on top to provide confidence measures for the model prediction.

\myparagraph{OOD selective prediction.}
~\cite{geifman2019selectivenet} proposes SelectiveNet that incorporates a selection head on the top of the base model, which is optimized with a selective loss to reject samples that the model is uncertain about. 
\cite{kamath-etal-2020-selective} trains a calibrator on top of an existing NLP model to generalize to unknown OOD data at test time.
Specifically, it trains the calibrator on a mixture of some held-out ID data and  `known' OOD data.
The final model is used for the evaluation of the unknown OOD data at test time.

\myparagraph{OOD detection.}
Earlier works~\cite{hendrycks17baseline} rely on the maximum class probability (MaxProb) to detect OOD samples.
\cite{liang2017enhancing} proposes ODIN that combines temperature scaling and image perturbation to achieve better separation in softmax scores for OOD and ID images.
Another line of work uses distance~\cite{lee2018simple} or energy~\cite{liu2020energy, lin2021mood, wang2021can} scores for OOD detection.
\cite{wang2022vim} introduces VIM that detects OOD samples by fusing the logits and feature information obtained from the model.
\cite{tian2014anomaly, bergman2020deep, sun2022knn} computes nearest-neighbor distances in the feature dimension to detect OOD data.

\myparagraph{Image OOD detection \& reliability.}
\cite{ovadia2019can} investigates the effect of dataset distribution shift on accuracy and calibration.
\cite{lakshminarayanan2017simple} uses deep ensembles to quantify uncertainty estimates of classification models.
\cite{gawlikowski2021survey, abdar2021review} extensively review of uncertainty estimation methods in deep learning literature.

\section{Selective VQA with ID and OOD Data}
In this section, we discuss the problem formulation of Selective VQA in \cref{sec:problemform}, and how we evaluate in the ID (in-distribution) scenario (\cref{sec:selectpredeval}) and in the  mixed ID+OOD (out-of-distribution) scenarios (\cref{sec:eval:OOD}).
\subsection{Problem Formulation}\label{sec:problemform}
The primary setting for VQA is to learn a function $f:\mathcal{Q,V}\mapsto\mathcal{A}$ to predict an answer $a\in\mathcal{A}$ to a question $q\in\mathcal{Q}$ about a given image $v\in\mathcal{V}$~\cite{antol2015vqa,goyal2017vqav2,gurari2018vizwiz}.
However, when exposing models to the real world they might encounter hard questions, OOD data points, or even adversarial questions by users and we cannot expect that models are able to answer all questions in these scenarios correctly. Therefore, we instead would like to identify inputs $x=(v,q) \in \mathcal{X}$ that models cannot correctly answer and abstain in those cases.
This is the setting of Selective Prediction~\cite{elyaniv2010foundations}, which has also recently been studied for ID VQA~\cite{whitehead2022reliable} and OOD text-only question answering \cite{kamath-etal-2020-selective}.
In this work, we advocate for this selective prediction setting for ID and OOD scenarios. We closely follow  the formalism introduced in \cite{whitehead2022reliable} for VQA, though it is very similar to setups outside of VQA (e.g., ``classification with a rejection option''~\cite{chow1970optimum,destefano00tsmc,geifman2019selectivenet,hanczar2008classification,pudil92icpr}, or ``selective prediction/classification''~\cite{elyaniv2010foundations,geifman2017selective}).
Specifically, the output space is extended to allow for an abstention option (denoted by $\emptyset$): $h:\mathcal{X}\mapsto\mathcal{A}\cup\{\emptyset\}$.
Such a \emph{Selective Model}  $h$ can be realized by decomposing $h$ into two functions, a VQA model $f$ and selection function $g:\mathcal{X} \mapsto \{0, 1\}$~\cite{elyaniv2010foundations,geifman2017selective,geifman2019selectivenet,whitehead2022reliable}:
\begin{equation}\label{eq:select}
    h(x) = (f,g)(x) = 
    \begin{cases}
    f(x)& \text{if}\  g(x) = 1, \\
    \emptyset & \text{if}\ g(x) = 0. \\
    \end{cases}
\end{equation}
For a given image-question pair $x=(v,q)$, the Selective VQA model $h$ only predicts an answer from the VQA model $f$ if the selection function $g$ decides that an answer should be given.
Otherwise, the Selective VQA model $h$ abstains. The selection function $g$ can be formulated  based on a function $g': \mathcal{X}\mapsto \mathbb{R}$ that scores the correctness of the model's prediction $f(x)$~\cite{geifman2019selectivenet,kamath-etal-2020-selective,whitehead2022reliable}, and a threshold $\gamma \in \mathbb{R}$.
Then, for a given $\gamma$, the model outputs the answer $f(x)$ if $g'(x)\geq\gamma$ and abstains otherwise.
Ideally, $g'$ should yield higher values if $f(x)$ is correct and lower if it is incorrect.
However, as we show in the experiments this is a hard task.

\begin{figure*}[t]
\includegraphics[width=\textwidth]{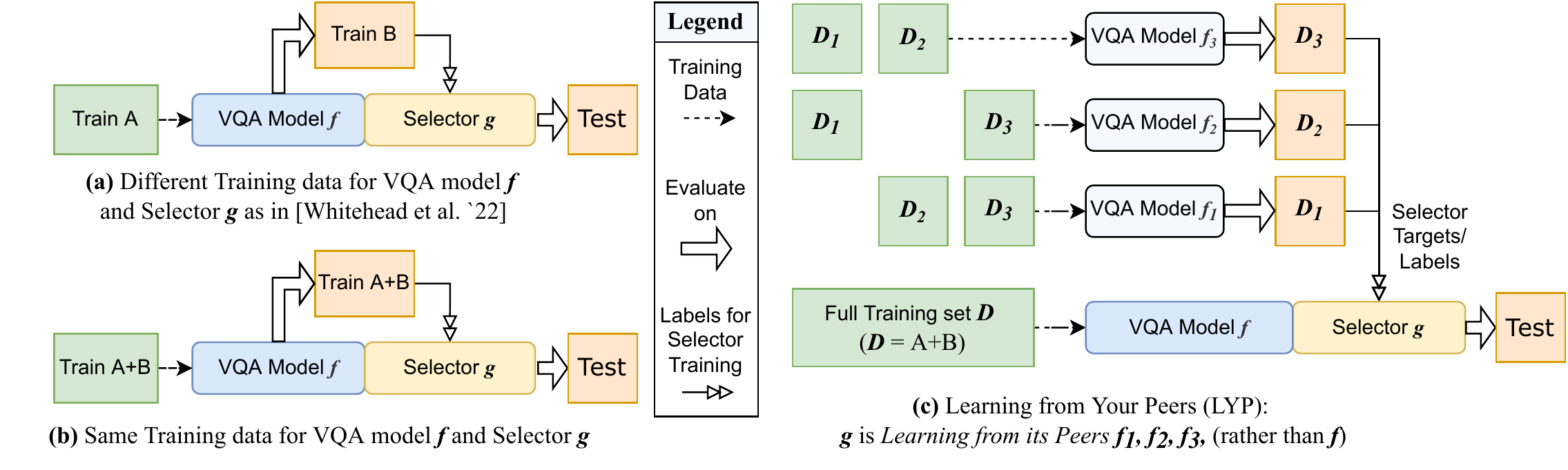}\vspace{-3mm}
   \caption{Comparison between \selector $g$ training procedures. (a) shows the one in \cite{whitehead2022reliable}, (c) shows our \ours. See \cref{sec:method} for details.}    \label{fig:method-comparison}
 \end{figure*}

\subsection{Evaluation}\label{sec:selectpredeval}

Beyond accuracy, we evaluate using the metrics designed for models with abstention options following~\cite{whitehead2022reliable}:

\myparagraph{Risk and coverage.}
For a dataset $D$, model $f$, and a selection function $g$, \emph{coverage} is the proportion of answered questions:
$$\mathcal{C}(g) = \frac{1}{|D|}\sum_{x \in D}g(x) ,$$
while \emph{risk} is the average error on the covered subset
$$\mathcal{R}(f, g) = \frac{
    \sum_{(x_i, y_i) \in D}(1 - \mathrm{Acc}(f(x_i), y_i)) \cdot g(x_i)
}{
    \mathcal{C}(g)
} ,$$
where $\mathrm{Acc}$ is VQA accuracy~\cite{antol2015vqa} and $y_i$ is the corresponding ground truth answer.
We measure the maximum coverage at a specific risk tolerance, denoted (\symcovatr), by determining the largest consecutive subset of questions that can be answered with at most $\mathcal{R}$ risk.
Further, we also compute the Area Under the Curve (AUC) for the risk-coverage curve \cite{kamath-etal-2020-selective} for a summary of performance across different coverage levels.

\myparagraph{Effective reliability $\Phi_c$.}
This metric is introduced in \cite{whitehead2022reliable} to better compare methods on the test set for a threshold selected on a validation set.
This is especially important for OOD, as thresholds for a certain risk level don't  generalize to the test scenario. $\Phi_c$ is a cost-based metric and jointly measures the reliability and effectiveness of selective models in a single metric.
It assigns a cost of $c$ to every wrong answer that the model outputs (i.e., does not abstain on):
\begin{equation}\label{eq:effectiveness}
        \Phi_c(x) = 
    \begin{cases}
    Acc(x)& \text{if}\  g(x) = 1\  \text{and}\  Acc(x) > 0,\\
    -c & \text{if}\ g(x) = 1\  \text{and}\  Acc(x) = 0,\\
    0 & \text{if}\ g(x) = 0.
    \end{cases}
\end{equation}
The total score is $\Phi_c =\frac{1}{|D|}\sum_{x \in D}\Phi_c(x)$, a mean over all samples $x$.
To compute this metric, we set the threshold $\gamma$ on a validation set to maximize $\Phi_c$.
Then, we use this threshold for abstention decisions on the test set.

\subsection{Evaluating with Mixed ID+OOD Data}
\label{sec:eval:OOD}

As previously mentioned, we want to explore the setting where models encounter mixtures of ID and OOD data.
More formally, we assume we are given $\mathcal{D}_{\mathrm{train}}$ and $\mathcal{D}_{\mathrm{test}}$ that are drawn from different distributions.
In our setting, to simulate a setting closer to a real-world use case, the test data is sampled from a mixture of ID and OOD data.
Similar to~\cite{kamath-etal-2020-selective}, we assume that our training data is drawn from $P_{\mathrm{src}}$ while our testing data is drawn from $P_{\mathrm{tgt}}$, where $P_{\mathrm{tgt}} = \alpha P_{\mathrm{src}} + (1 - \alpha) P_{\mathrm{unk}}$.
Here, $P_{\mathrm{unk}}$ is an unknown distribution different from $P_{\mathrm{src}}$ from which we obtain our OOD examples.
We obtain different mixtures of data by varying $\alpha$ and evaluate models across these using the metrics discussed in \cref{sec:selectpredeval}.
Different from prior work in NLP~\cite{kamath-etal-2020-selective}, we assume we \emph{do not} have access to known OOD data for training, meaning all models must be trained and thresholds must be chosen on ID data.
However, we do compare to this setting in our experiments.

\newcommand{\midsplitcline}{\arrayrulecolor{lightgray}\cmidrule(r){2-2}\cmidrule(lr){3-5} \cmidrule(lr){6-6}  \cmidrule(lr){7-9} \cmidrule(lr){10-10} \cmidrule(lr){11-13}\arrayrulecolor{black}}

\begin{table*}[t]
\centering
\footnotesize
\begin{tabular}{lclclrrrrrrrr}
\toprule
\multicolumn{2}{c}{VQA Model $f$} & \multicolumn{3}{c}{Selection func. $g$}  & \multirow{2}{*}{ \grey{Acc $\uparrow$}} & \multicolumn{3}{c}{\symcovatr in $\% \uparrow$} & \multirow{2}{*}{AUC $\downarrow$} & \multirow{2}{*}{\effrel{1}} & \multirow{2}{*}{\effrel{10}} & \multirow{2}{*}{\effrel{100}}  \\
\cmidrule(lr){1-2} \cmidrule(lr){3-5} \cmidrule(lr){7-9}
Name & Train Set &  Name & Train Set & Targets& & \covatr{1} & \covatr{5} & \covatr{10} \\
\midrule
\multirow{5}{*}{CLIP-ViL} & \multirow{2}{*}{\train}
& MaxProb & - & - \hspace{10pt}\cite{whitehead2022reliable}  & \grey{69.98} & 4.97 &33.79 &53.62 &10.92 &54.67 &21.40 &1.32 \\
&& Selector & \dev & Self \cite{whitehead2022reliable} & \grey{69.98} & 15.79 & 37.79 & 55.65 & 10.21 & 55.44 & 25.82 & 8.74 \\
\midsplitcline
& \multirow{3}{*}{\traindev} 
& MaxProb  & - & - & \grey{70.72} & 5.54 & 34.84 & 55.04 & 10.49 & 55.93 & 22.81 & 2.59 \\
&& Selector & \traindev & Self & \grey{70.72} & 6.45 & 34.26 & 56.07 & 10.48 & 56.07 & 22.99 & 2.39 \\
&& Selector & \traindev & \ours & \grey{70.72}  & \textbf{18.40} & \textbf{38.65} & \textbf{57.40} & \textbf{9.76} & \textbf{56.53} & \textbf{26.45} & \textbf{9.74} \\ 
\midrule
\multirow{5}{*}{\ofabase} & \multirow{2}{*}{\train} 
& MaxProb & - & - & \grey{74.87} & 3.45 & 45.60 & 66.61 & 7.99 & 62.52 & 30.57 & 6.81 \\ && Selector & \dev & Self & \grey{74.87} & 23.78 & 49.16 & 69.00 & 7.32 & 63.03 & 34.39 & 12.53 \\
\midsplitcline
& \multirow{3}{*}{\traindev}
& MaxProb & - & - & \grey{75.18} & 14.88 & 46.15 & 67.51 & 7.79 & 63.04 & 30.13 & 7.29\\ && Selector & \traindev &  Self & \grey{75.18} & 26.64 & 50.80 & 69.56 & 7.10 & 63.66 & 34.92 & 12.92 \\ 
&& Selector & \traindev & \ours & \grey{75.18} & \textbf{27.71} & \textbf{51.64} & \textbf{70.20} & \textbf{6.98} & \textbf{63.88} & \textbf{36.29} & \textbf{16.30}\\
\midrule
\multirow{5}{*}{\ofalarge} & \multirow{2}{*}{\train} 
& MaxProb  & - & - & \grey{77.53} & 20.57 & 53.99 & 75.18 & 6.42 & 66.68 & 36.12 & 8.21 \\
&& Selector & \dev & Self & \grey{77.53} & 30.86 & 58.05 & 76.65 & 5.81 & 67.34 & 41.43 & 17.58 \\
\midsplitcline
& \multirow{3}{*}{\traindev} 
& MaxProb & - & - &  \grey{77.79} & 16.31 & 53.83 & 75.27 & 6.43 & 66.96 & 36.06 & 6.29 \\
&& Selector & \traindev &  Self &  \grey{77.79} & 31.47 & 58.80 & 77.14 & 5.69 & 67.82 & 41.43 & 16.08 \\
&& Selector & \traindev & \ours & \grey{77.79} & \textbf{32.92} & \textbf{59.43} & \textbf{77.52} & \textbf{5.60} & \textbf{68.02} & \textbf{42.83} & \textbf{18.78} \\
\bottomrule
\end{tabular}\caption{
Risk-coverage metrics and effective reliability on ID data (i.e., \vqatwo test split from~\cite{whitehead2022reliable}).
}
\label{tab:main-results}
\end{table*}

\section{\ours: \fullours}\label{sec:method}

Prior work has established training a selection function (or \selector) $g$ to predict the correctness of the outputs of a model $f$~\cite{geifman2019selectivenet,kamath-etal-2020-selective,whitehead2022reliable} as a method for selective prediction.
As in \cite{whitehead2022reliable}, our \selector $g$ learns to predict the VQA Accuracy of $f$.
One option is to train $f$ on one part of the training data (Train \train) and  $g$ on a different, typically smaller, part (Train \dev), as shown in \cref{fig:method-comparison}(a). Having separate training data for $g$ can be crucial since if $f$ has overfit to the training data, then training $g$ on that same data will lead $g$ to a solution that doesn't generalize well (e.g., always answering). We show some of these drawbacks in our experiments with observations similar to findings on stacked generalization~\cite{wolpert1992stacked}.
However, withholding data from training $f$ could reduce the overall performance of $f$, as it does not allow $f$ to learn from this data.
Likewise, $g$ is unable to learn from the training data for $f$.
This motivates training both $f$ and $g$ on the same data, e.g., as done in \cite{geifman2019selectivenet} (shown in \cref{fig:method-comparison}(b)).

We propose a simple yet effective approach, called \fullours (\ours), for training $g$ that allows both $f$ and $g$ to utilize all the available training data.
Inspired by work on collective outliers~\cite{karamcheti2021mindyouroutliers} and improving worst group performance~\cite{liu2021justtraintwice}, our approach aims to identify examples in the training data that are difficult to generalize to, for a given architecture and learning procedure.
In particular, we want to provide more signal to $g$ about which examples in the training data may not be generalizable and likely should be abstained on, despite the VQA model's potential ability to fit these examples during training.

Shown in \cref{fig:method-comparison}(c), we first partition our full training set $\mathcal{D}$ into $N$ disjoint subsets ($\mathcal{D}=$ Train \train + Train \dev).
For our VQA setting, we create our partitions by ensuring no images overlap between them.
Next, we train $N$ different models on combinations of the subsets in leave-one-out manner: we create a training set $\mathcal{D}^{*}_{n} = \mathcal{D} \setminus \mathcal{D}_n$ and train a VQA model $f_n$ on $\mathcal{D}^{*}_n$.
Once we have trained $f_n$, we use it to make predictions on $\mathcal{D}_n$, which it has not seen during its training.
We use the ground truth annotations for $\mathcal{D}_n$ to obtain VQA accuracy for each prediction, which we treat as a label for the correctness of each prediction.
After performing this operation for $n=1,...,N$, we can union the partitions to obtain an updated training set $\mathcal{D}^{\mathrm{sel}}$ that additionally has correctness labels for each example $(x^{(n)}_i, y^{(n)}_i, f_n(x^{(n)}_i), \xi^{(n)}_i)$ for $(x^{(n)}_i, y^{(n)}_i) \in \mathcal{D}_n$, where $\xi^{(n)}_i = \mathrm{Acc}(f_n(x^{(n)}_i), y^{(n)}_i)$.

We train our VQA model $f$ on all of $\mathcal{D}$ and then, with the obtained correctness labels, we train our Selector $g$ on top of $f$ using the full $\mathcal{D}^{\mathrm{sel}}$ dataset.
For training $g$, we follow~\cite{whitehead2022reliable} and optimize it using a regression objective with the correctness labels as the target.
Note, our setup is similar to that of \cite{whitehead2022reliable} in that we use a regression objective, but, importantly, the source of our targets is not the model $f$ itself but, rather, the subset models $\{f_n\}^{N}_{n=1}$ (i.e., the \emph{peers} of $f$).
The idea behind this is that if a model trained on the remainder of the training data $\mathcal{D}^{*}_n$ cannot generalize to an example in $\mathcal{D}_n$, then that may be a challenging example that $g$ should choose to abstain on as the model $f$ is unlikely to generalize reliably to such an example at test time, even if it has fit it during training.
Essentially, these correctness labels may provide a signal for which examples are difficult and might require abstention \emph{more generally} rather than with respect to a specific model as in prior work~\cite{whitehead2022reliable}.
Moreover, we show in our experiments that this allows $f$ and $g$ to learn from the entire training data, which can boost overall accuracy as well as abstention performance.
Our method requires training $N$ models, which can be done in parallel, but, unlike ensembling, we have a single model for inference.

\begin{table*}[t!]
\centering
\footnotesize
\begin{tabular}{lclclrrrrrrrr}
\toprule
\multicolumn{2}{c}{VQA Model $f$} & \multicolumn{3}{c}{Selection function $g$}    & \multirow{2}{*}{ \grey{Acc $\uparrow$}} & \multicolumn{3}{c}{$\mathcal{C} @ \mathcal{R}$ in $\% \uparrow$} & \multirow{2}{*}{ AUC $\downarrow$} & \multirow{2}{*}{\effrel{1}} & \multirow{2}{*}{\effrel{10}} & \multirow{2}{*}{\effrel{100}} \\
\cmidrule(lr){1-2} \cmidrule(lr){3-5} \cmidrule(lr){7-9}
Name & Train Set&  Name & Train Set & Targets& & \covatr{1} & \covatr{5} & \covatr{10} \\
\midrule  
\multirow{5}{*}{ CLIP-ViL } 
    & \multirow{2}{*}{\train}
& MaxProb & - & - \hspace{10pt}\cite{whitehead2022reliable} & \grey{66.35} & 0.00 & 24.16 & 43.53 & 13.55 & 49.12 & 14.39 & -4.64 \\
&& Selector & \dev & Self~\cite{whitehead2022reliable} &	\grey{66.35} & 12.69 & 31.12 & 46.96 & 12.47 & 50.36 & 20.15 & 5.22 \\
\midsplitcline
& \multirow{3}{*}{\traindev} 
& MaxProb  & - & - & \grey{67.12} & 2.60 & 26.13 & 45.25 & 12.97 & 50.49 & 16.59 & -0.93 \\
&& Selector & \traindev &  Self & \grey{67.12} & 2.97 & 26.70 & 46.19 & 12.80 & 50.89 & 18.19 & -0.65 \\
&& Selector & \traindev & \ours & \grey{67.12} & \textbf{15.22} & \textbf{32.58} & \textbf{49.18} & \textbf{11.90} & \textbf{51.43} & \textbf{22.09} & \textbf{7.12} \\
\midrule
\multirow{5}{*}{\ofabase} & \multirow{2}{*}{\train} 
& MaxProb & - & - &  \grey{71.59} & 0.01 & 36.07 & 56.49 & 10.10 & 57.49 & 23.15 & -0.34 \\
&& Selector & \dev & Self & \grey{71.59} & 18.32 & 41.48 & 59.74 & 9.19 & 57.97 & 27.22 & 9.09 \\
\midsplitcline
& \multirow{3}{*}{\traindev} 
& MaxProb & - & - & \grey{72.00} & 1.74 & 37.02 & 57.57 & 9.78 & 58.11 & 22.09 & 0.53 \\
&& Selector & \traindev &  Self & \grey{72.00} & 19.72 & 42.70 & 60.84 & 8.90 & 58.90 & 28.05 & 2.88 \\
&& Selector & \traindev & \ours & \grey{72.00} & \textbf{21.58} & \textbf{44.09} & \textbf{61.69} & \textbf{8.74} & \textbf{59.11} & \textbf{28.79} & \textbf{10.88} \\
\midrule
\multirow{5}{*}{\ofalarge} & \multirow{2}{*}{\train} 
& MaxProb & - & - & \grey{74.56} & 4.76 & 44.53 & 66.06 & 8.21 & 61.90 & 28.20 & 0.21 \\
&& Selector & \dev & Self & \grey{74.56} & 23.53 & 50.17 & 68.76 & 7.33 & 62.96 & 34.43 & 9.88 \\
\midsplitcline
& \multirow{3}{*}{\traindev} 
& MaxProb & - & - & \grey{74.79} & 1.30 & 43.70 & 65.95 & 8.26 & 62.24 & 27.09 & -2.46  \\
&& Selector & \traindev & Self & \grey{74.79} & 22.68 & 50.27 & 69.27 & 7.32 & 63.03 & 33.50 & 4.92 \\
&& Selector & \traindev & \ours & \grey{74.79} & \textbf{25.38} & \textbf{51.07} & \textbf{69.74} & \textbf{7.17} & \textbf{63.41} & \textbf{34.85} & \textbf{10.34} \\
\bottomrule
\end{tabular}
\caption{Results on the mixed ID/OOD scenario composed of 90\% \vqatwo and 10\% \advqa examples.
}
\label{tab:main-advqa}
\end{table*}

\section{Experiments}
\subsection{Setup}\label{sec:expsetup}

\myparagraph{Data.} We require both ID and OOD data that has annotations available for evaluation.
Therefore, we utilize the splits of the \vqatwo dataset~\cite{goyal2017vqav2} made available by~\cite{whitehead2022reliable} as our ID data.
The entire \vqatwo train set (call it split \textbf{\train}) is used for training VQA models ($f$).
Meanwhile, the \vqatwo validation set is split into 3 parts: 86k examples (40\%) for training selection functions $g$ (call it split \textbf{\dev}); 22k examples (10\%) for validating models; 106k examples (50\%) as a test split for evaluating full selective models $h=(f,g)$.
\ours does not require different sets for training $f$ and $g$, so we train them both with the combination of A and B (\textbf{\traindev}).
For OOD data, we use \advqa~\cite{sheng2021advqa}, which is an adversarial dataset constructed by asking human annotators to create questions that are difficult to answer for existing VQA models trained on \vqatwo.
The images in \advqa and \vqatwo overlap with each other, so we only use images from \advqa that appear in the test split.
While \advqa is not OOD in terms of the images, one can still consider this as adversarial, OOD since the questions are designed to fall outside the training distribution of \vqatwo.
This is similar to other OOD VQA datasets like VQA-CP~\cite{agrawal2018vqacp}, VQA-CE~\cite{dancette2021beyond}, or other VQA generalization benchmarks~\cite{agrawal2022rethinking,whitehead2021separating}.
However, for our setting, we create mixtures of \vqatwo and \advqa to serve as our evaluation data, where each mixture contains different amounts of ID/OOD data.

\myparagraph{VQA models.}
We use two different VQA architectures: \textbf{CLIP-ViL}~\cite{shen2021clipvil}, which is an ensemble of MCAN~\cite{yu2019deep} and MoVie~\cite{nguyen2021movie} with a CLIP~\cite{radford2021learning} image encoder, and the recent \textbf{OFA} model~\cite{wang2022ofa}, which is a transformer encoder-decoder model that performs multiple tasks and achieves state-of-the-art accuracy on \vqatwo.
For OFA, we explore 2 different sizes of the model: Base and Large.
CLIP-ViL is a strong VQA model that treats VQA as a classification task over a large set of answers~\cite{teney2018tips}, while OFA is a large-scale pre-trained model that treats VQA as a generative task\footnote{While OFA is a generative model, it uses a trie-based decoding method for VQA that restricts the generated sequences to an answer vocabulary, as opposed to open-ended generation~\cite{wang2022ofa}.}.

\myparagraph{Selection functions.}
We explore \textbf{MaxProb}~\cite{geifman2017selective,guo2017calibration,hendrycks17baseline,kamath-etal-2020-selective,whitehead2022reliable} as a baseline as it is a natural comparison to the VQA model out-of-the-box since the confidence scores are simply the output probabilities of the model.
We also employ the \textbf{Selector} developed by \cite{whitehead2022reliable} as it attains the strongest performance for selective VQA.
Selector is a two-layer MLP that takes in a combination of image, question, multimodal, and answer representations from the VQA model in order to predict a confidence score.
We apply \ours to train Selector and compare to training with the original approach in \cite{whitehead2022reliable} that utilizes held-out data.
For each approach, we set a threshold on the output confidence scores to make abstention decisions (\cref{sec:problemform}). Unless specified, we use by default $N=10$ disjoint subsets to partition our \traindev data.

All results for the strongest VQA model, \ofalarge, are averaged over 5 runs, while all other results are single runs.
More experimental details are in the appendix.

\subsection{In-Distribution Experiments}
\label{sec:experiments:ID}

We first experiment with only in-distribution data to compare with prior work.
Discussed in \cref{sec:problemform}, we evaluate using maximum coverage at different risk levels (\symcovatr), AUC for the risk-coverage curve, and effective reliability at different costs (\symeffrel).
We also present accuracy to give an idea of the question-answering performance of each model.

\myparagraph{ID performance consistently improves with \ours.}
\cref{tab:main-results} shows that across all model architectures the top scores are achieved using \ours.
For instance, we see improvements in \covatr{1} over both \maxprob(\traindev) and \selector(\dev) with OFA-Large of 16.61\% and 2.06\%, respectively.
Likewise, \effrel{100} increases with \ours by 12.49 and 1.20 over \maxprob(\traindev) and \selector(\dev), respectively, for \ofalarge.
The improvements are sustained at higher risk levels and lower costs (e.g., +0.63\% \covatr{10} for \selector with \ours for \clipvil compared to \selector trained on held-out data).
These observations hold across each model we experiment with on ID data.
Lastly, we see that all \selector models outperform all \maxprob models on every metric, just as in \cite{whitehead2022reliable}.

\myparagraph{\ours helps VQA models and \selector learn from the same data.}
We see that training \selector and \clipvil on the same data (\traindev) performs poorly, achieving \symcovatr and \symeffrel similar to its \maxprob counterpart.
Conversely, the \ofa models and \selector are able to be trained on the same data and reap the benefits of training on more data.
We conjecture that this is due to the overfitting issue discussed in \cref{sec:method} as \clipvil has a training accuracy of 87.40\% whereas, e.g., \ofabase has a training accuracy of 82.92\% while also having higher accuracy on the test split.
However, when using \ours, \clipvil and \selector can be trained on the same data and improve beyond the model of \cite{whitehead2022reliable} by, e.g., 2.61\% \covatr{1}.
Further, although training on the same data can be done for the \ofa models and \selector, it does not perform quite as well as when \ours is used.
For example, with \ofabase, training both the VQA model and \selector on \traindev has \covatr{1} of 26.64\% compared to 23.78\% when the VQA model is trained on \train and \selector is trained on \dev.
Meanwhile, using \ours with \ofabase attains 27.71\% \covatr{1}.
Overall, these results suggest \ours helps better utilize the training data with Selector, improving ID performance.

\subsection{OOD Evaluation}
\label{sec:experiments:OOD}
For our OOD evaluations, we build mixed datasets comprised of 10\%, 33\%, 50\%, and 66\% OOD examples.
All mixtures contain 5K examples from \advqa as OOD examples, and the rest are randomly sampled from the ID \vqatwo test split.
We report the results on the 10\% OOD mixture in \cref{tab:main-advqa}.
More details, results (e.g., on other mixtures), and qualitative examples are in the appendix.

\myparagraph{\maxprob can be overconfident on OOD data.}
Across all models, we see that \maxprob has $<$5\% \covatr{1} and \symeffrel scores $<$1.
This suggests that \maxprob can be overconfident on OOD examples, on which the model is more likely to be incorrect.
While improving the VQA accuracy of the model improves MaxProb performance, training a \selector still remains the most effective approach and consistently.

\begin{figure}[t]
    \centering
    \includegraphics[width=0.8\linewidth]{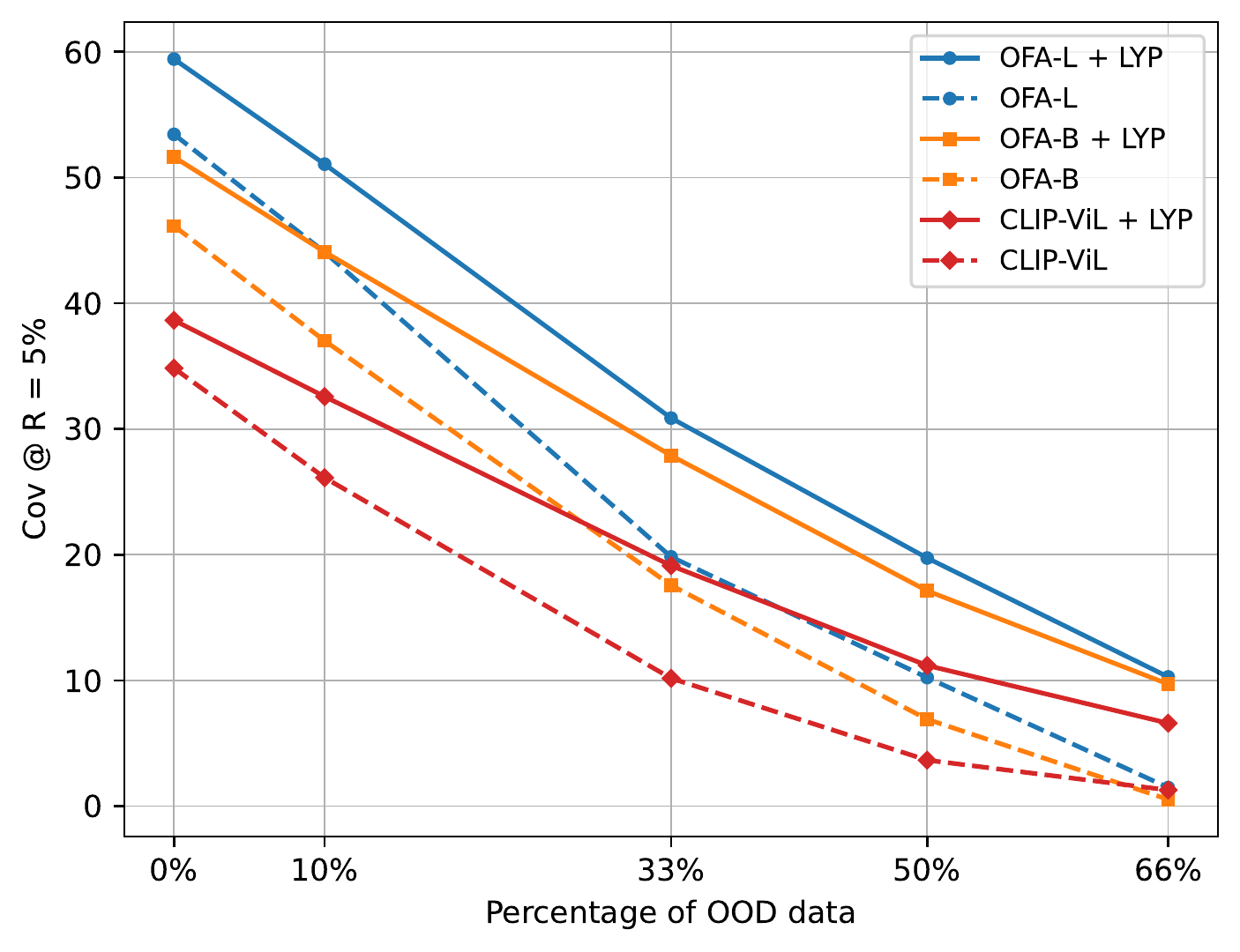}
    \caption{\covatr{5} for various mixtures of \vqatwo + \advqa. OFA-L stands for Large, OFA-B for Base. Baseline is MaxProb.}
    \label{fig:ood-proportion-r5}
    \end{figure}

\myparagraph{\ours maintains improvements over other methods in the 90\%/10\% setting.}
Similar to the pure ID setting, \ours continues to outperform other methods on the 90\%/10\% mixed setting as shown in \cref{tab:main-advqa}.
Although, we see decreases in all metrics across each of the different methods, demonstrating the challenge of this task even with just 10\% OOD data.

\begin{figure}[t]
    \centering
    \includegraphics[width=0.8\linewidth]{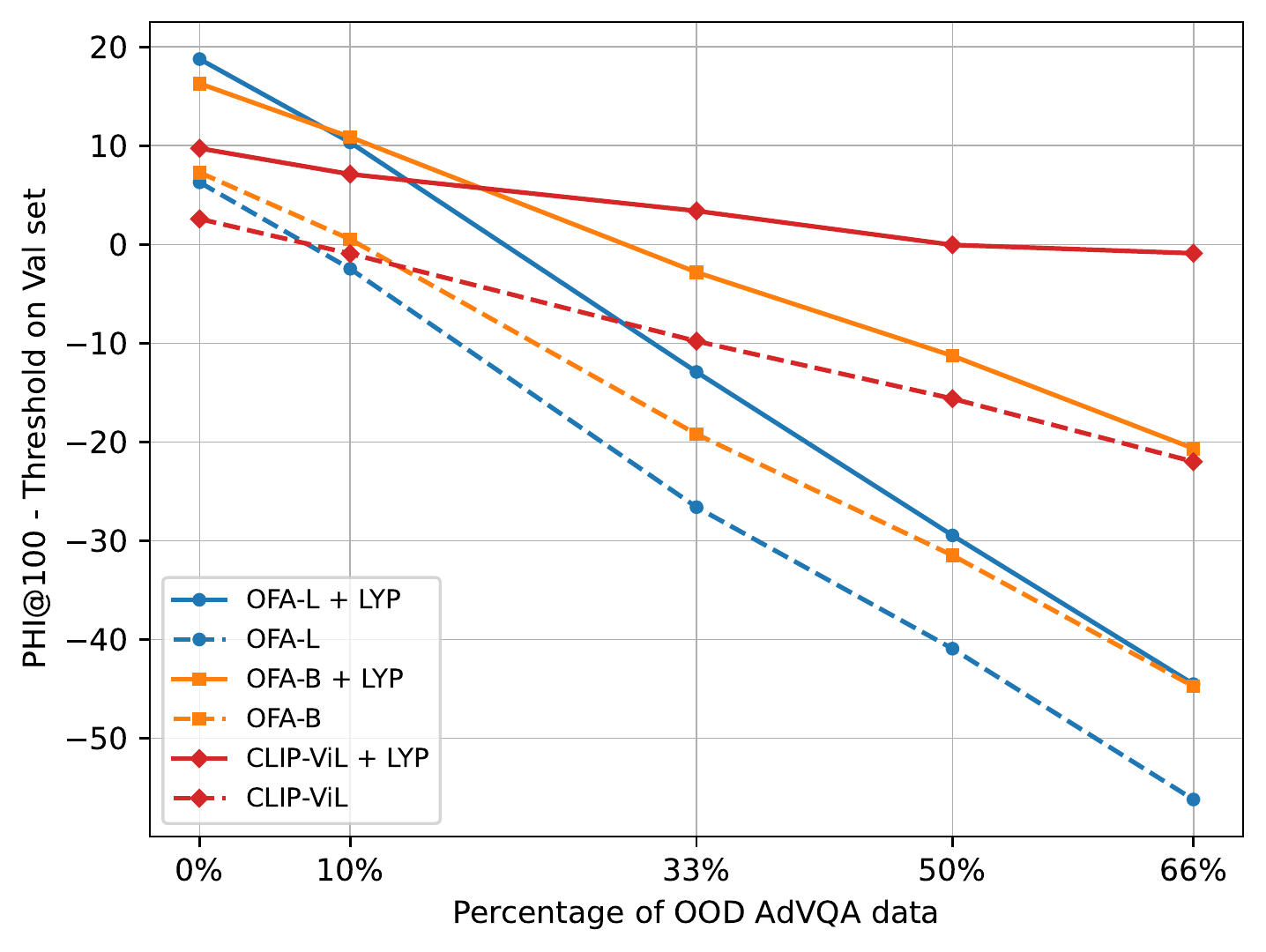}
    \caption{\effrel{100} for various mixtures of \vqatwo + \advqa. OFA-L stands for Large, OFA-B for Base. Baseline is MaxProb.}
    \label{fig:ood-proportion-phi}
        \end{figure}

\myparagraph{The more OOD data, the more challenging.}
We display \covatr{5} and \effrel{100} for the various mixtures of ID/OOD data in \cref{fig:ood-proportion-r5,fig:ood-proportion-phi}.
For both plots, we show each of the three models with a \ours-trained Selector versus the performance of MaxProb, each trained on the full \traindev data.
Across all OOD levels, \ours largely outperforms the baseline for all three models, for both \covatr{5} and \effrel{100} metrics. 
However, we observe that performances degrade quickly with a high OOD level.
At the highest level (i.e., 33.3\%/66.7\% ID/OOD), all Maxprob models have $<$2\% \covatr{5}, while LYP has around 10\% coverage. For \effrel{100}, most models are below zero.
We see that scaling alone is not sufficient to ensure high performances: while \ofalarge (\maxprob) has good performances on ID data, and is above \ofabase + \ours, this is no longer true with OOD data.
Our LYP Selector is effective at mitigating this loss in performance on OOD data.
However, for \ofalarge, we note that the LYP-trained Selector is not always better than the Selector trained with held-out data for higher OOD levels.
We discuss this and a mitigation strategy in the appendix.
Combining the observations in  \cref{fig:ood-proportion-r5,fig:ood-proportion-phi}, we see the potential performance that models could achieve, based on \symcovatr which is irrespective of the threshold chosen, versus the realized performance when choosing a threshold as one would do in practice, shown by \effrel{100}.
This shows that more work is needed to help generalize to such OOD data.

\myparagraph{OOD detection features do not necessarily help.}
Inspired by~\cite{fisch2022calibrated}, we train \selector with out-of-distribution detection scores computed with KNN~\cite{sun2022knn} or SSD~\cite{sehwag2021ssd} as added features.
We find that these features do not bring significant improvements to our evaluation metrics (\cref{tab:ood-detection}).
More details about those experiments can be found in the appendix.

\begin{table}[t]
    \centering
    \footnotesize
    \begin{tabular}{llcccccc}
     & Acc. & \covatr{1} & \covatr{5} & \covatr{10}  & AUC \\
   \midrule
   \selector       & 71.25 & 19.05 & 41.83 & 59.55 & 9.29 \\
   \selector + KNN & 71.25 & 19.92 & 41.78 & 59.75 & 9.27 \\
   \selector + SSD & 71.25 & 18.99 & 41.90 & 59.27 & 9.27 
           \end{tabular}
    \caption{OOD Detection baselines. Scores are reported on the mixed ID/OOD data composed of 90\% \vqatwo\ / 10\% \advqa.}
    \label{tab:ood-detection}
\end{table}

\myparagraph{Augmenting Selector training with known OOD data also does not improve.}
As discussed in \cref{sec:eval:OOD}, we also try training \selector on the \dev set, along with some known OOD datasets similar to \cite{kamath-etal-2020-selective}.
This may help learn to discard hard examples which are very far from its training distribution.
For this experiment, we use the training sets of OK-VQA~\cite{marino2019okvqa}, which has the same image distribution but a different question distribution, and of VizWiz~\cite{gurari2018vizwiz}, which has both image and question distribution shifts compared to \vqatwo.
We see in \cref{tab:ood-exposure} that this method is not very successful at improving selective prediciton in this OOD evaluation setting.
Contrary to the findings of~\cite{kamath-etal-2020-selective} for text-only question answering, on the Selective VQA task, adding this known OOD data during training decreases the performance of our Selector on unknown OOD data at test time.
Overall, it appears that more traditional approaches for handling OOD examples may have difficulty generalizing to this multimodal setting.

\begin{table}[t]
    \centering
    \footnotesize
    \begin{tabular}{llcccccc}
        Model & Selector & \covatr{1} & \covatr{5} & \covatr{10}  & AUC \\
        \midrule
        \multicolumn{6}{c}{90\% \vqatwo, 10\% \advqa} \\
        \midrule
        \train & \dev & 19.00 & 41.64 & 58.97 & 9.34   \\
        \train & \dev + OOD & 18.48 & 41.08 & 59.40 & 9.36\\
        \midrule
        \multicolumn{6}{c}{50\% \vqatwo, 50\% \advqa} \\
        \midrule
        \train & \dev & 2.68 & 15.98 & 26.72 & 18.97  \\
        \train & \dev + OOD & 2.56 & 14.93 & 26.82 & 19.08 \\

            \end{tabular}
    \caption{Results with exposure to known OOD data for \ofabase.}
    \label{tab:ood-exposure}
\end{table}

\subsection{Further Analysis}
In addition to the following, we have more qualitative results, analysis, and evaluation on other tasks (visual entailment~\cite{xie2019snlive}) and datasets (VizWiz~\cite{gurari2018vizwiz}) in the appendix.

\myparagraph{Different numbers of splits/peers for \ours.}
We ablate the number of splits/peer models $N$ of the training data $\mathcal{D}$ with \ofabase.
Ideally, the peer models in \ours should have predictions and failure modes similar to the full model, which suggests that more peer models may be better (i.e., each peer model is trained on more data).
We see in \cref{tab:num-splits} that for \ofabase the number of peers has a small impact on the final results.
However, we also find that the difference in accuracy of \ofabase fine-tuned on 50\% vs 100\% of the training data is small (74.03\% vs 75.18\%).
Therefore, the difference in signal from the labels of 2 vs 10 peers may be similar.
For VQA models with less pre-training, this difference may be greater and more models may be needed.
This suggests that the training requirements for \ours can be reduced while maintaining strong performance for large pre-trained models.

\begin{table}[t]
\footnotesize
\centering
    \begin{tabular}{ccccc}
        \midrule
        N  & \covatr{1} & \covatr{5} & \covatr{10}  & AUC \\
        \midrule
                        10 & \textbf{27.71 }& \textbf{51.64} & \textbf{70.20} & \textbf{6.98} \\
        2  & 27.64 & 51.24 & 70.12 & 7.01 \\
                                                        \midrule
    \end{tabular}
\caption{Varying the number of splits $N$ for \ours. Results are reported on the ID test split for \ofabase, trained on \traindev, with a selector trained on \traindev.}
    \label{tab:num-splits}
\end{table}

\myparagraph{Effect of training data size.}
We show in \cref{tab:data-percent} that the amount of data used for the \selector training is an important factor for its performance.
Note that the Train \dev set, used by ~\cite{whitehead2022reliable} to train their selector, has 86K examples, which is $\sim$15\% of the full Train \traindev.
The additional data, labeled with \ours, helps \selector generalize better to test examples.

\begin{table}[t]
    \centering
    \footnotesize
    \begin{tabular}{cccccc}
       \% of \traindev & \covatr{1} & \covatr{5} & \covatr{10}  & AUC \\
       \midrule
       100 & \textbf{27.71 }& \textbf{51.64} & 70.20 & \textbf{6.98} \\
       75  & 27.48 & 51.11 & \textbf{70.26} & 7.01 \\
       50  & 26.84 & 51.04 & 70.04 & 7.06 \\
       25  & 26.03 & 50.15 & 69.65 & 7.16 \\
       10  & 23.30 & 47.97 & 68.03 & 7.44 \\
       5   & 22.62 & 46.10 & 66.10 & 7.71 \\
    \end{tabular}
    \caption{Varying the amount of training data for the \selector with \ours. The model is \ofabase and results are on the ID test split.}
        \label{tab:data-percent}
\end{table}

\myparagraph{Impact of scaling on selective prediction.}
\cref{tab:scaling} shows results for three OFA sizes: Medium, Base, and Large.
We see that larger models, in addition to having a much higher accuracy on the testing set, have much better ID selective prediction performance when paired with a trained Selector.

\begin{table}[t]
    \centering
    \footnotesize
    \begin{tabular}{llcccccc}
       Model & Method & Acc. & \covatr{1} & \covatr{5} & \covatr{10}  & AUC \\
    \toprule
       Med. & \maxprob & 71.30 &  5.08 & 37.56 & 56.85 & 9.95  \\
       Base & \maxprob & 74.70 & 3.45  & 45.60 & 66.61 & 7.99 \\
       Large & \maxprob  & 77.79 & 20.57 & 53.99 & 75.18 & 6.42 \\
       \midrule
       Med. & \ours & 71.30 & 19.69 & 41.28 & 59.60 &  9.17 \\ 
       Base & \ours   & 74.70 & 27.71 & 51.64 & 70.20 & 6.98 \\
       Large & \ours  & 77.79 & 32.92  & 59.43 & 77.52 & 5.60 \\
           \end{tabular}
    \caption{Scaling results for OFA Medium (93M params), Base (180M params), and Large (470M params) on the ID test split.}
        \label{tab:scaling}
\end{table}

\section{Conclusions}

This is the first work to explore Selective Visual Question Answering in the realistic, and challenging, mixed ID+OOD scenario, where a model is exposed to samples from both the training distribution and also out-of-distribution (OOD) examples.
We find that out-of-the-box, state-of-the-art VQA models~\cite{shen2021clipvil,wang2022ofa} largely fail on this task at a low risk of error (e.g., 1\%).
When training a multimodal Selector~\cite{whitehead2022reliable} models significantly improve their abstention decisions, matching observations in the in-distribution (ID) scenario.
However, a limitation of the multimodal Selector training is that it requires splitting the training data between the VQA model training and the Selector training to avoid over-fitting on the training data.
In this work, we address this with our approach \emph{\fullours} (\ours), which allows us to train both the VQA model and the Selector on the full training data. 
We find that in the ID scenario as well as the mixed scenario of 90\%/10\% ID/OOD data, \ours consistently performs best across all VQA models and metrics, improving over baselines and prior work.
Our best result doubles the \covatr{1} over prior work~\cite{whitehead2022reliable}.
Overall, all models still have difficulties recognizing when they cannot answer OOD examples correctly and thus decrease in performance when the percentage of OOD data increases.
Interestingly, we observe that the better a VQA model is ID, the more it loses if it has to also generalize the threshold for abstention from ID to OOD (as measured by Effective Reliability \symeffrel).
Thus, major challenges remain, both for improving the generalizing abilities of VQA models to OOD examples (i.e., answering OOD questions correctly) as well as identifying examples that the model cannot answer, whether they are in- or out-of-distribution.

\section*{Acknowledgements}

From the Sorbonne Universit\'e side, this effort was partly supported by ANR grant VISADEEP (ANR-20-CHIA-0022). This work was granted access to the HPC resources of IDRIS under the allocation 2022-AD011011588R2 made by GENCI.

{\small
\bibliographystyle{ieee_fullname}
\bibliography{egbib}
}

\clearpage
\appendix

\section*{Index}
\begin{description}
\item[\cref{sec:appendix:OOD-OKVQA}] shows additional ablations for adding known OOD data, confirming the findings in the main paper.
    \item[\cref{sec:appendix:OODresults}] provides the result tables for all the ID/OOD mixtures, including the larger percentages of OOD examples.
    \item[\cref{sec:jointraing}] compares our staged training setup to jointly training VQA model and selector.
    \item[\cref{sec:app:setup}] has the details of the experimental setup, including model details and dataset splits.
    
        \item[\cref{sec:ood:features}] has the details of the OOD Detection features, which we use in the experiments in Table 3 of the main paper.
    \item[\cref{sec:thresholdgeneralization}] offers a closer look at the difficulty in estimating the abstention threshold on in-domain data, when OOD data is present at test time.
    \item[\cref{sec:vizwiz}] presents experiments using VizWiz~\cite{gurari2018vizwiz} as the source of OOD data.
    \item[\cref{sec:snlive}] presents selective prediction experiments on another multimodal task: SNLI-VE~\cite{xie2019snlive}.
    \item[\cref{sec:app:qualitative}] illustrates qualitative results in \cref{fig:advqa:abstain,fig:advqa:correct,fig:advqa:incorrect}.
\end{description}

\section{Training with Known OOD Data}
\label{sec:appendix:OOD-OKVQA}
To further understand the usefulness of additional OOD data for Selector training in our multimodal setting as suggested in \cite{kamath-etal-2020-selective} for text-only NLP tasks, we provide an additional ablation: In \cref{tab:ood-exposure-supp} for the ID/OOD training setup, we train only on the \dev set + the OK-VQA training set, i.e. without VizWiz data (third line in each section of the table). The OK-VQA is more similar to AdVQA compared to VizWiz. However, we observe similar results compared to using both OOD datasets: Additional Known OOD does not consistently improve the results over the baseline (i.e. most identical to the selector setup in \cite{whitehead2022reliable}), especially for low risk, the model with OK-VQA does not perform well. Alternative use of such known OOD data in the multimodal setting is out of scope for this work, but it is an interesting avenue for future work to study how to potentially better exploit such data.

\begin{table}[ht]
    \centering
    \footnotesize
    \begin{tabular}{llcccc}
    \toprule
    \multicolumn{2}{l}{Train Set} & \\  
         $f$ & Selector $g$ & \covatr{1} & \covatr{5} & \covatr{10}  & AUC \\
        \midrule
        \multicolumn{6}{c}{90\% \vqatwo, 10\% \advqa} \\
        \midrule
        \train & \dev & \textbf{19.00} & \underline{41.64} & 58.97 & \underline{9.34}   \\
        \train & \dev + OOD & \underline{18.48} & 41.08 & \underline{59.40} & 9.36\\
        \train & \dev + OK-VQA & 18.38 & \textbf{42.33} & \textbf{59.80} & \textbf{9.17} \\
        \midrule
        \multicolumn{6}{c}{50\% \vqatwo, 50\% \advqa} \\
        \midrule
        \train & \dev & \textbf{2.68} & \textbf{15.98} & \underline{26.72} & \underline{18.97}  \\
        \train & \dev + OOD & \underline{2.56} & 14.93 & \textbf{26.82} & 19.08 \\
        \train & \dev + OK-VQA & 1.73 & \underline{15.37} & 26.33 & \textbf{18.86} \\
    \bottomrule
    \end{tabular}
    \caption{Results with exposure to known OOD examples for \ofabase. OOD = OK-VQA + VizWiz. \textbf{Bold} denotes best and \underline{underline} is second best per table section.}
    \label{tab:ood-exposure-supp}
\end{table}

\section{Additional OOD Results}
\label{sec:appendix:OODresults}

\begin{figure}[ht]
    \centering
    \includegraphics[width=\linewidth]{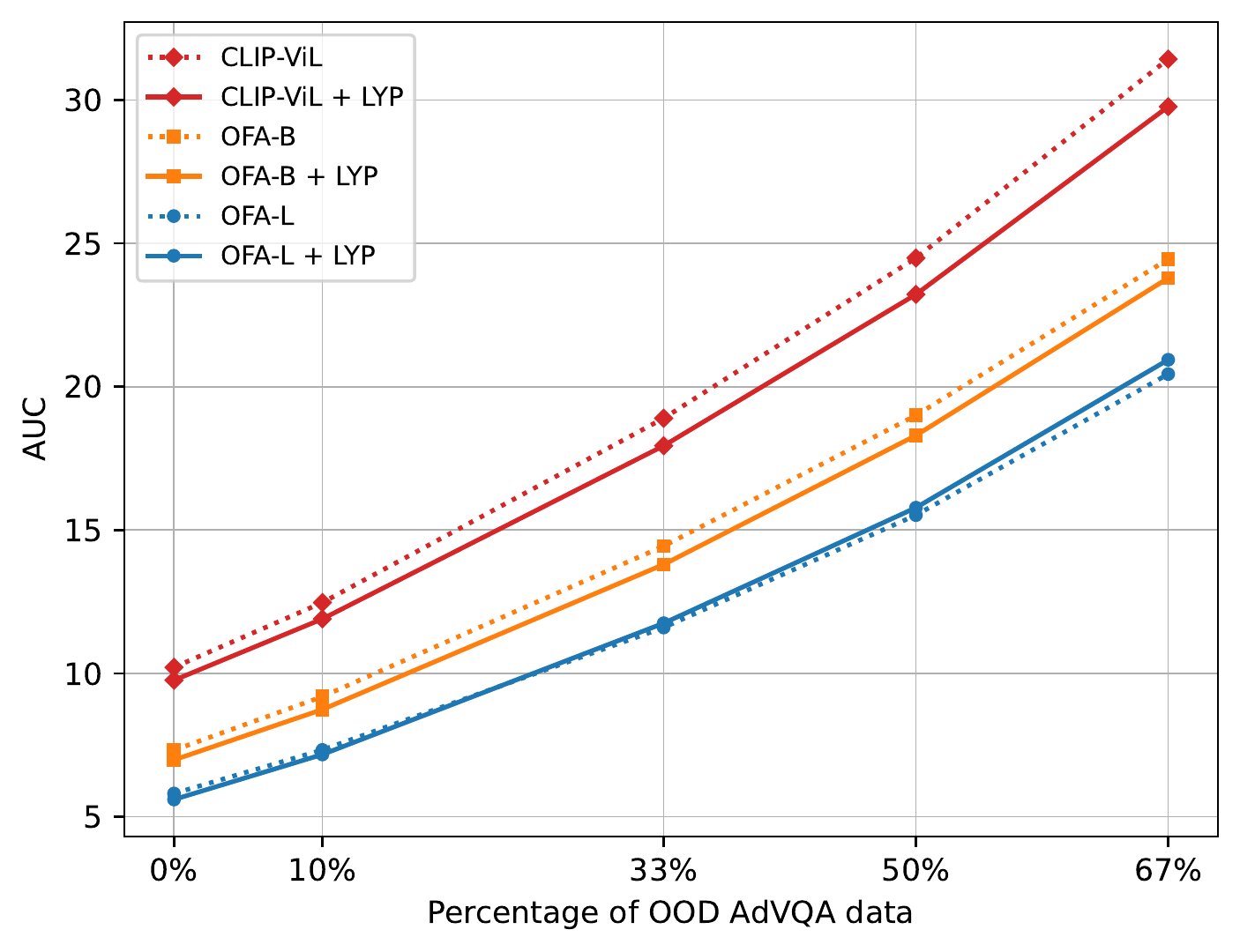}
    \caption{AUC for various mixtures of \vqa + AdVQA. Note: Lower is better for AUC. The baselines for each model is MaxProb.}
    \label{fig:ood-proportion-auc}
    \end{figure}

We show the AUC for our models on various mixtures of ID/OOD data in \cref{fig:ood-proportion-auc}.
Overall, our method consistently improves AUC over the baseline, for the three models (note lower is better for AUC).

In \cref{tab:supp:main-results,tab:supp:main-advqa,tab:main-advqa-33,tab:main-advqa-50,tab:main-advqa-66}, we present the results for our experiments on the all OOD mixtures of \vqatwo and \advqa~\cite{sheng2021advqa}.
While we on average see that the \selector models and \selector + \ours perform better than the corresponding baselines models out-of-the-box (\maxprob), all models degrade dramatically if there is a high percentage of OOD data in the test mixture, especially for low risk (\covatr{1}) or high cost of error (\effrel{100}).
Especially if we look at the realistic scenario where the threshold is chosen on the validation set and used at test time (as for \effrel{100}), we notice that the scores of all methods drop below 0 with 33.3\% or more OOD data.
This can be seen in the last column of \cref{tab:main-advqa-33,tab:main-advqa-50,tab:main-advqa-66}.
These results demonstrate that these thresholds can be overconfident on OOD examples, which leads to poor abstention decisions such that the cost of the models' incorrect outputs outweighs the gains of the correct ones.
Future work is needed to improve such OOD generalization and recognizing samples that cannot reliably be answered in this challenging setup, which this work provides a new and interesting test setup for.

\subsection{Alternative \ours Strategy}
\label{sec:supp:LYPASelfB}
\begin{figure}
    \centering
    \includegraphics[width=\linewidth]{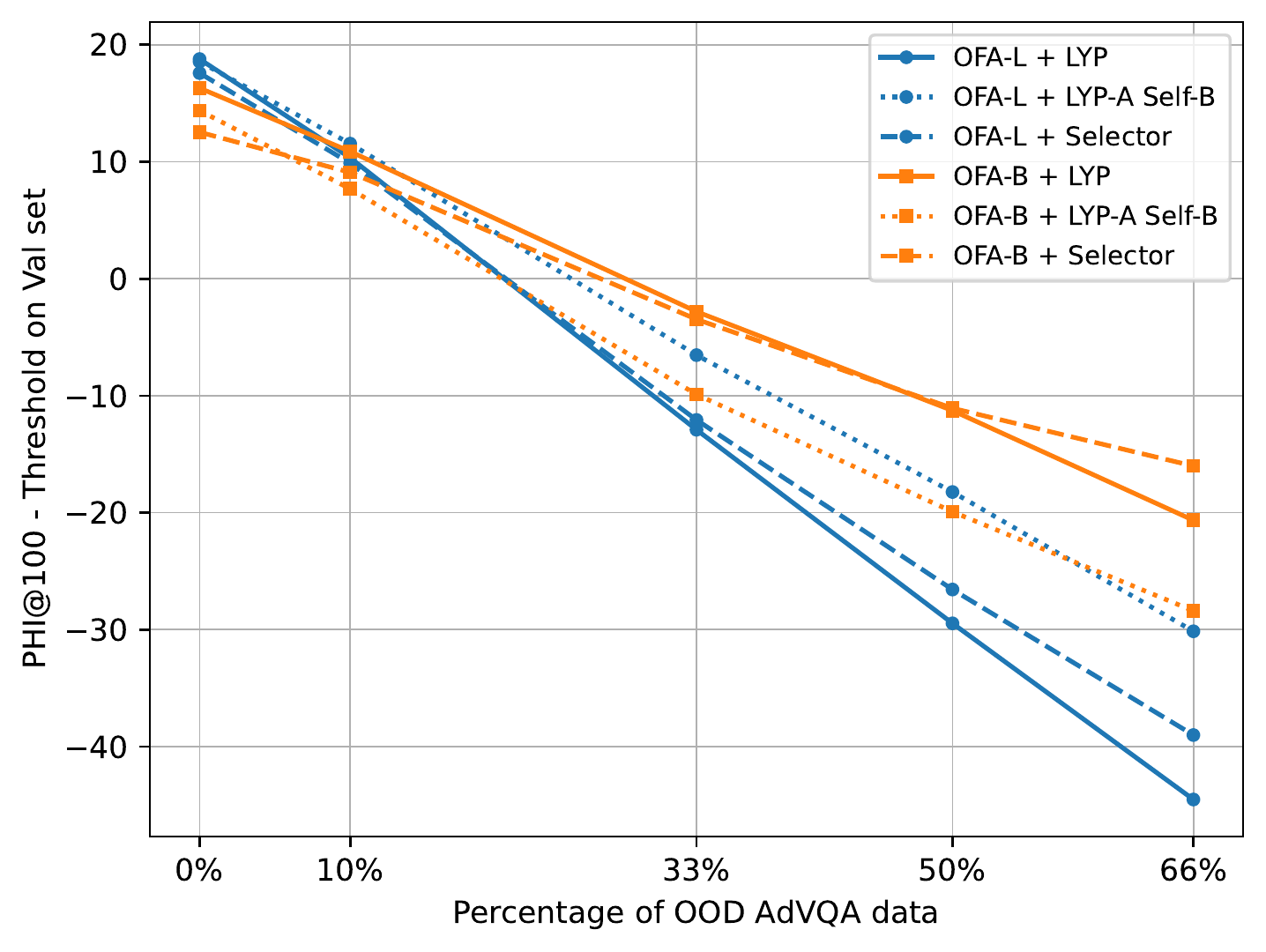}
    \caption{\effrel{100} scores for OFA-L and OFA-B models. The Selector is trained on \dev, based on the VQA model trained on \train.  We also show the alternative selector training strategy ``LYP-A Self-B''.}
    \label{fig:phi100:lyp-a-self-b}
\end{figure}

\begin{figure}
    \centering
    \includegraphics[width=\linewidth]{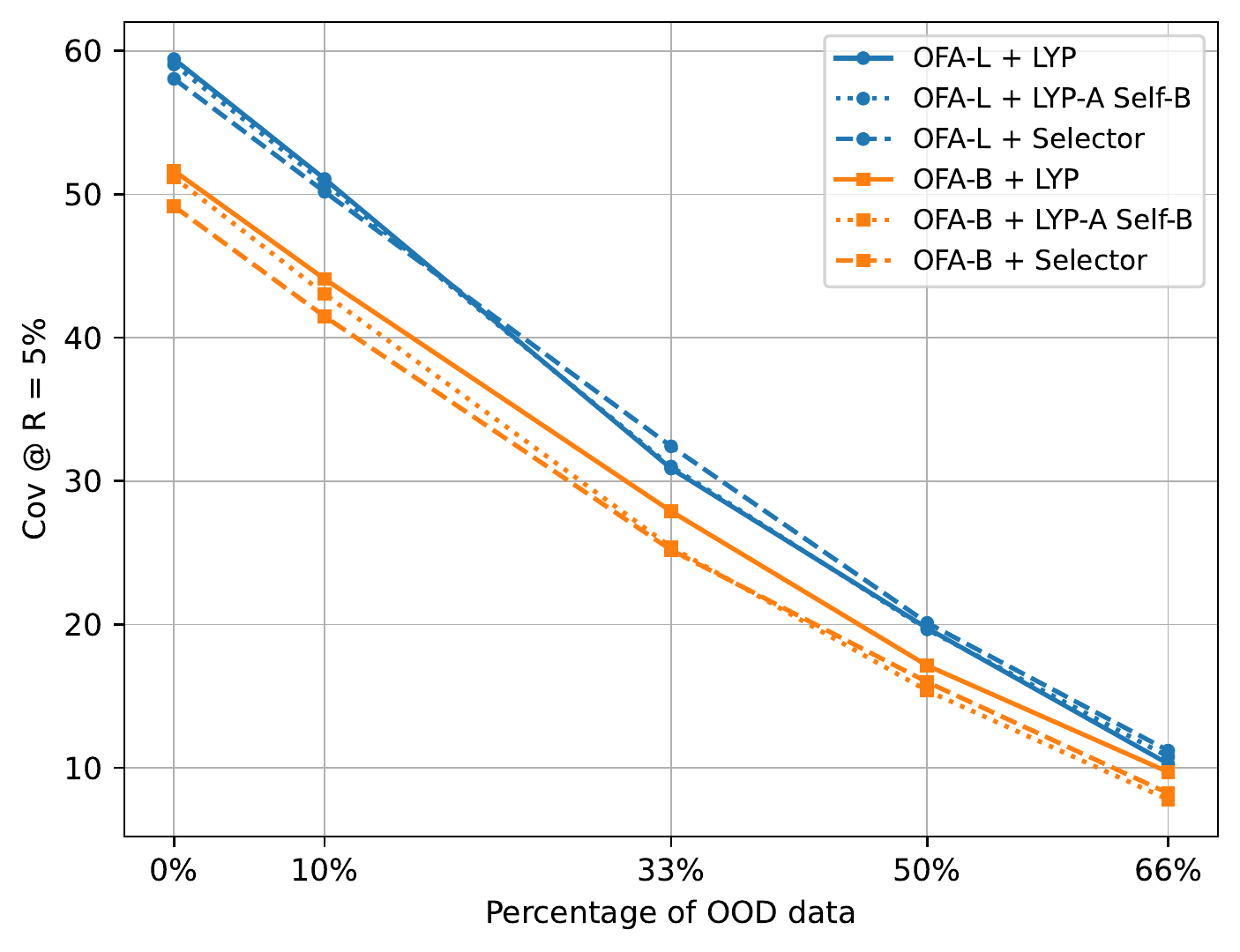}
    \caption{\covatr{5} scores for OFA-L and OFA-B models. The Selector is trained on \dev, based on the VQA model trained on \train. We also show the alternative selector training strategy ``LYP-A Self-B''.}
    \label{fig:covat5:lyp-a-self-b}
\end{figure}

As mentioned in the main paper, there are cases where LYP does not perform quite as well as the baseline \selector that is trained on held-out data.
This happens with OFA-Large on high OOD levels, particularly with the \effrel{100} metric, which involves generalizing a confidence threshold chosen on ID data to test time where both ID and OOD data are present.
This is shown in \cref{fig:phi100:lyp-a-self-b,fig:covat5:lyp-a-self-b}, where at higher percentages of OOD, \ofalarge with \ours (+\ours) can have lower \covatr{5} and \effrel{100} than the baseline \selector trained on held-out data (+\selector).

We propose a potential mitigation strategy for such cases.
First, we use the VQA model trained only on the \train subset, instead of the one trained on the full \traindev like in LYP.
Then, we train a Selector on the full \traindev data, using the following strategy: we use \ours only on the \train subset and use the model's own labels on the \dev subset.
This allows the Selector to be trained partly on some data that was unseen during the VQA model's training, with real confidence labels.
This potentially helps the selector capture the model's real uncertainty.
We call this strategy \textbf{LYP-A Self-B} and report it in all \cref{tab:supp:main-results,tab:supp:main-advqa,tab:main-advqa-33,tab:main-advqa-50,tab:main-advqa-66}.

\cref{fig:phi100:lyp-a-self-b} illustrates the effect of this method on \ofabase and \ofalarge for the \effrel{100} metric.
We show that using this LYP-A Self-B strategy improves the \effrel{100} scores significantly for \ofalarge, surpassing both the baselines and the standard LYP.
On \ofabase, however, the base \selector and \ours perform similarly while LYP-A Self-B under-performs them.
Therefore, it appears that this method should not be applied in all cases but might help to improve the results when \ours is less effective (e.g., because the VQA model overfits too much on the training set while not benefiting sufficiently from the additional training data in B).

\section{Jointly Training \ofa and \selector}
\label{sec:jointraing}
Discussed in \cref{sec:app:setup}, for training \selector, we follow a staged procedure~\cite{whitehead2022reliable}:
The VQA model is first trained until convergence on the VQA task.
Then, the weights are frozen, \selector is added to the model, and \selector is learned on top of the frozen model.

Since we are able to train \ofa and \selector on the same data, a natural comparison to make is between the staged training procedure we use and joint training (i.e., simultaneously optimizing the VQA model and \selector), similar to~\cite{geifman2019selectivenet}.
We experiment with joint training by summing their losses.
We perform this on \ofabase, training both \ofabase and \selector with the full \traindev data.
We also experiment with first joint training \ofabase and \selector until \ofabase has converged for the VQA task, freezing \ofabase, and continuing to fine-tune \selector on \traindev.

The results in \cref{tab:joint-supp} illustrate that joint training decreases the overall performance of the \selector.
All metrics yield worse performance with joint training alone, though the gap shrinks when freezing the VQA model and continuing to fine-tune \selector.
This is despite the fact that the overall VQA accuracy remains roughly the same with or without joint training.
We conjecture that the reason for this may be that joint training creates a somewhat non-stationary optimization problem for \selector.
Specifically, the VQA model's representations and VQA accuracy are changing throughout training.
This means that the statistics of the inputs and training targets for \selector (see \cref{sec:app:setup}) are changing, which may make optimizing \selector more difficult.
Other techniques may be needed in order to properly optimize the VQA model and \selector together.

\begin{table}[t]
    \centering
    \footnotesize
    \begin{tabular}{lccccc}
    \toprule
        Training & \grey{Acc} &\covatr{1} & \covatr{5} & \covatr{10}  & AUC \\
        \midrule
        \multicolumn{6}{c}{ID (100\% \vqatwo)} \\
        \midrule
        joint & \grey{75.08} & 16.04 & 42.78 & 65.91 & 8.11 \\
        joint+FT & \grey{75.08} & 24.42 & 50.01 & 69.20 & 7.21 \\
        staged & \grey{75.18} & \textbf{26.64} & \textbf{50.80} & \textbf{69.56} & \textbf{7.10} \\
        \midrule
        \multicolumn{6}{c}{90\% \vqatwo, 10\% \advqa} \\
        \midrule
        joint & \grey{71.97} & 10.74 & 34.61 & 53.81 & 10.12 \\
        joint+FT & \grey{71.97} & 18.17 & 42.44 & 60.50 & 8.98 \\
        staged & \grey{72.00} & \textbf{19.72} & \textbf{42.70} & \textbf{60.84} & \textbf{8.90} \\
    \bottomrule
    \end{tabular}
    \caption{Comparison of joint and staged training of \ofabase and \selector. FT indicates that \selector is further fine-tuned after \ofabase converges on the VQA training objective. All models are trained on \traindev.}
    \label{tab:joint-supp}
\end{table}

\section{Experimental Setups}\label{sec:app:setup}

\subsection{Models}

\subsubsection{\ours Peer Models}
Our \ours approach requires training \emph{peer} models to label the training data for the full \selector.
For all \ours peer models, we simply follow the corresponding VQA model training settings.
Once trained, we run inference on the respective held-out sets for each peer to obtain labels.

\subsubsection{CLIP-ViL}

We use the implementations for \maxprob and \selector provided by~\cite{whitehead2022reliable}.\footnote{\url{https://github.com/facebookresearch/reliable_vqa}}
For the CLIP-ViL \maxprob and \selector models trained on held-out data (i.e., Train \dev), which exactly match the setup of \cite{whitehead2022reliable}, we use the model weights given by the authors as well.
Note that the available model weights are for a single run, whereas the results in \cite{whitehead2022reliable} are averaged over ten runs, so there are some variations in scores between those reported in this work and \cite{whitehead2022reliable}.
Additionally, we compare to the results in the arXiv version of~\cite{whitehead2022reliable} as this has the most up-to-date results (see the appendix of \cite{whitehead2022reliable}).
For the remaining CLIP-ViL models, we train them following the provided hyperparameters and settings.
We refer readers to~\cite{whitehead2022reliable} for details.

\subsubsection{OFA}
\ofa first processes the image using a convolutional network~\cite{he2016deep} to obtain a set of visual representations $\tilde{V}$.
Likewise, the question is tokenized and converted to a sequence of question token embeddings $\tilde{Q}$.
Then, the visual features are flattened into a sequence and concatenated with the question token embedding sequence.
This entire sequence is given as input to an encoder-decoder transformer model~\cite{vaswani2017attention} to predict the answers.
The encoder produces multimodal representations of the image tokens $\{v_i\}_{i=1}^{|\tilde{V}|}$ and question tokens $\{q_j\}_{j=1}^{|\tilde{Q}|}$.
The encoded tokens are used as input to cross-attention layers in the transformer decoder at each decoding step.
The decoder generates output token representations $\{o_l\}_{l=1}^{L}$ for an answer of $L$ tokens.
These token representations can be fed to a linear layer to give the output logits over the token vocabulary.
We use beam search to decode the answers.

We fine-tune OFA from the pre-trained checkpoints provided by the authors of~\cite{wang2022ofa}.\footnote{\url{https://github.com/OFA-Sys/OFA}}
We follow the hyperparameters from the original paper for fine-tuning.
In the following, we detail the setup for the selection functions:

\myparagraph{\maxprob.} Since \ofa is a sequence-to-sequence model that generates answers token-by-token, for the \maxprob baseline, we use the joint probability of each answer token as the confidence value, similar to common decoding algorithms like beam search.

\myparagraph{\selector.}
We largely replicate the same \selector architecture and training as~\cite{whitehead2022reliable} (i.e., two-layer MLP), but with some slight differences.
We remove the non-linear projection (or one-layer MLP) for each input representation.
We also use slightly different input representations:
First, we max-pool the encoder image ($v_i$) and question ($q_i$) token representations to obtain a single representation for each set of representations.
Then, we extract the probability of the predicted answer $p$, which is the joint probability of each answer token.
Finally, we extract the first output token embedding $o_1$ that is used to predict the first answer token.
We concatenate these representations and feed this as input to the \selector.

\begin{table}[t]
    \centering
    \begin{tabular}{@{}l@{}rr@{}}
    \toprule
        & \ofabase & \ofalarge \\
    \midrule
       Batch Size  &  256 & 512 \\
       Learning Rate & 1e-4 &  4e-4 \\
       LR warmup & no & no \\
       LR-decay (linear) & -1e-10/step & -1e-10/step \\
       Optimizer & Adam & Adam\\
       Optimizer Beta & (0.9,0.999) & (0.9,0.999) \\
       Gradient clipping & 1.0 & 1.0 \\
       Selector Dropout & 0.1 & 0.1 \\
       Main model dropout & 0.1 & 0.1 \\
       Image size & 480 & 480 \\ 
       \bottomrule
    \end{tabular}
    \caption{Hyperparameters for \selector Training on top of \ofa}
    \label{tab:hyperparams}
\end{table}

\myparagraph{Training \selector with \ofa.}
We report the training parameters in \cref{tab:hyperparams}.
We first train the VQA model as discussed above, freeze the VQA model, and then train \selector on top of this frozen model, following~\cite{whitehead2022reliable}.
We train for a maximum number of 32 epochs and perform early-stopping on the \val split (\cref{tab:dataset-size}) using the AUC metric.
We keep the dropout in the main model during the selector training, as we found this improved performance of the selector.

\subsection{Dataset Splits}

\label{sec:datasetup}

\subsubsection{In-Distribution Splits}
We follow \cite{whitehead2022reliable} and use the splits provided in the official implementation.
We detail the splits again in \cref{tab:dataset-size}.
Note, in our work we repurpose the ``Dev'' set from \cite{whitehead2022reliable} for our Train B split.
No images (or question-answer annotations) are shared between splits.

\newcommand{\colseplocal}{\ \ \ \ }
\begin{table}[t]
    \centering
    \footnotesize
    \begin{tabular}{@{}l@{\colseplocal}l@{\colseplocal}l@{\colseplocal}r@{\colseplocal}r@{\colseplocal}r@{}}
    \toprule
     Split & Usage & Source &$\% \mathrm{src}$ & $\# \mathrm{I}$ & $\# \mathrm{Q}$  \\     \midrule
     Train A & Train $f$,$g$ & \vqa train & 100\% & 82,783 & 443,757 \\          Train B & Train $f,g$ & \vqa val  &40\% & 16,202 & 86,138 \\      Val & Validate $f,g$ & \vqa val  &10\% & 4,050 & 21,878 \\      Test & Test $h=(f,g)$ &\vqa val  & 50\% & 20,252 & 106,338 \\     \bottomrule
    \end{tabular}
    \caption{Size of the splits of \vqatwo from  \cite{whitehead2022reliable}. Note, the ``Usage'' is the setting for the full model (\traindev). Some models are trained on subsets (e.g., just \train) as specified in the corresponding tables.}
    \label{tab:dataset-size}
\end{table}

\subsubsection{ID/OOD Mixtures}
We use \advqa as our source of OOD data.
As discussed, \advqa is an adversarial dataset where human annotators intentionally ask questions that state-of-the-art models trained on \vqatwo answer incorrectly.
The images in \advqa come from \cite{lin2014coco}, as do \vqatwo.
However, we consider this as OOD since the questions are adversarial in nature and contain distribution shifts meant to induce errors for models trained on \vqatwo.

In our work, we create mixtures of ID/OOD examples for our evaluations.
To form our mixtures, we first discard all \advqa images that overlap with the \traindev train set.
This leaves 5,008 AdVQA examples.
For each setting, we randomly sample examples from the ID \test split (\cref{tab:dataset-size}) to create the desired OOD proportion: 45K for 10\% OOD, 10K for 33\% OOD, 5K for 50\% OOD and 2.5K for 66\% OOD.

\section{OOD Detection Features}\label{sec:ood:features}
In Table 3 of the main paper, we experiment with OOD detection features as additional input to the selector, inspired by~\cite{fisch2022calibrated}.
To compute those metrics, we use the representations from the encoder of \ofa.
We average the output question tokens $q_i$ and the image tokens $v_i$, which respectively yield $\bar{q}$ and $\bar{v}$.
We compute OOD detection features for each representation with respect to the training data.
The computed features are as follows:

\myparagraph{$k$NN~\cite{sun2022knn}.}
Given an input example, we compute the cosine distance to the $k$ nearest neighbors in the training data.
This distance is used as an OOD score: higher scores signify more ``in-distribution'' examples, while lower scores signify ``out-of-distribution''.
We use the efficient vector-search library FAISS~\cite{johnson2019faiss} to compute the distances and identify the $k$ closest points.
We experimented with various numbers of neighbors from 1 to 1000 and found no significant improvements for any value.
We also experimented with using the distance to \textit{correct} and \textit{incorrect} neighbors, to align the distances to our task of selective prediction.

\myparagraph{SSD~\cite{sehwag2021ssd}.}
SSD~\cite{sehwag2021ssd} is a parametric OOD-detection method that first builds $k$ clusters in feature space and then fits a multivariate normal distribution for each of the $k$ ensembles of features.
For a new example, the Mahalanobis distance~\cite{lee2018simple} to this normal distribution is used as an OOD score.
Note that for a classification task, the labels might be used as clusters, but we prefer to use a cluster-based algorithm, as the VQA answers do not represent a coherent ensemble of image or question concepts.
We experimented with various numbers of clusters in the range of $[1, 1000]$, and saw no improvements.

For these OOD detection features, we give them as additional inputs to the \selector to provide a signal for whether a given example is ID or OOD.

\section{Threshold Generalization}\label{sec:thresholdgeneralization}
In this section, we investigate threshold generalization.
All previous tables reported numbers on ``maximum coverage'' at risk $\mathcal{R}$.
This metric is irrespective of the threshold chosen as it solves for the coverage that satisfies a given risk level.
In a real-world setting, the threshold would need to be fixed once using a validation set and then used at test time.
We already evaluate this setting of evaluating the optimal threshold on the validation set for the cost-based metric \symeffrel in the main paper.
In contrast to \symeffrel, which allows comparing a single number, for risk and coverage, choosing a threshold on a validation set leads to changes in coverage \emph{and} risk, making it difficult to compare two methods.
Still, in this section, we evaluate how the threshold generalizes to ID and OOD settings.

\begin{figure}[t]
    \centering
    \includegraphics[width=\linewidth]{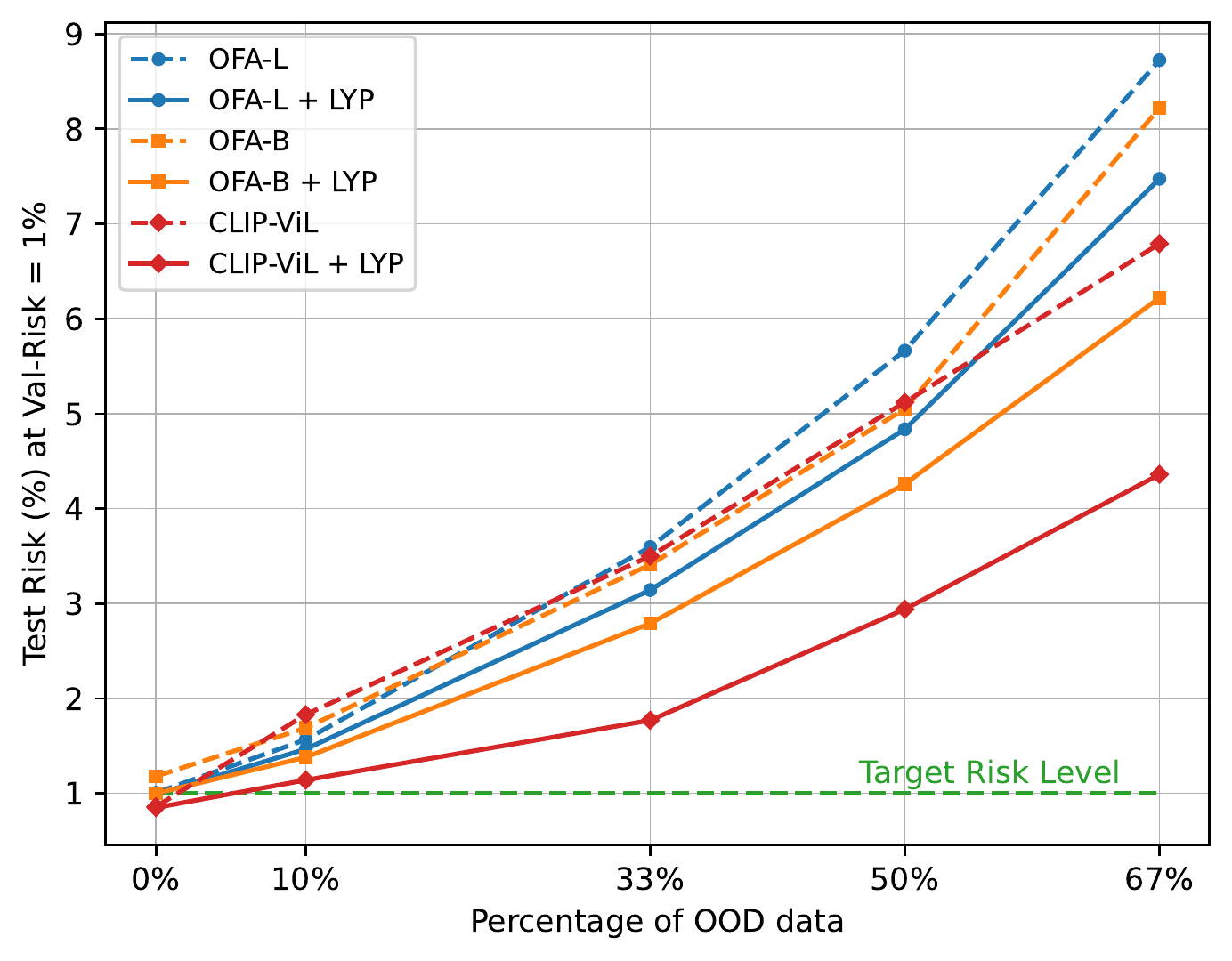}
    \caption{Risk at various percentages of OOD when the threshold is optimized on the validation set for maximum coverage, with a target risk level of 1\%. The baseline for each model is MaxProb.}
    \label{fig:ood-proportion-risk}
    \end{figure}

\myparagraph{Our method improves risk generalization over out-of-the-box \maxprob.}
In \cref{fig:ood-proportion-risk}, we show the test risk on various ID/OOD mixtures with a threshold set on the ID validation split of \vqatwo for a target risk of 1\%.
We see that \ours (solid line) consistently improves the generalization of risk over the MaxProb baseline: The curves corresponding to \ours are closer to the 1\% target risk level compared to MaxProb.

\myparagraph{Risk generalization is limited for OOD data.}
While we observe reasonable good risk generalization for ID, the generalization is really limited for larger percentages of OOD data.

\myparagraph{\clipvil is the best model for risk generalization.}
We see that all variants of CLIP-ViL outperform their corresponding methods on OFA-B and OFA-L.
Note that the associated coverages are lower for the same risk level, thus CLIP-ViL is not the best method overall. This is somewhat surprising, as \cite{kadavath2022llmknow} found that larger language models were better calibrated on NLP tasks.

Full results are available in \cref{tab:generalization:ID} and \cref{tab:generalization:90} for our in-distribution testing set and our mixed setting with 90\% of \vqa and 10\% of \advqa examples.

\section{VizWiz as OOD Data Source}\label{sec:vizwiz}

We show in \cref{tab:vizwiz,tab:vizwiz3} results for our LYP method on another OOD dataset: VizWiz \cite{gurari2018vizwiz}. This dataset is much more different from the original VQA v2 than AdVQA: the image and questions were collected by visually impaired users using a smartphone. Therefore, this makes it easier for models to discriminate between VQA and VizWiz examples.  We see that overall, the coverages are much higher for these setups than with AdVQA. We also see that our LYP method is very efficient to improve the results over the regular Selector model setup from \cite{whitehead2022reliable}.

\begin{table}[t]
\centering
\footnotesize
\setlength{\tabcolsep}{2.3pt}
\begin{tabular}{lccccccc}
&& &\multicolumn{5}{c}{10\% OOD} \\
\cmidrule(l){4-8} 
Model & $f$ train & $g$ train & AUC & \covatr{1} & \covatr{5} & \covatr{10} & \grey{Acc.} \\
\midrule
MaxProb  & A & -- & 10.61 & 9.44 & 37.4 & 54.30 & \grey{69.02} \\
Selector & A & B  & 9.14 &  25.76 & 46.57 & 62.07 & \grey{69.02} \\
Selector + LYP & AB & AB & \textbf{5.79} & \textbf{36.50} & \textbf{59.61} & \textbf{76.20} &\grey{76.12} \\
\end{tabular}
\caption{OFA-Base results with 90\% \textbf{VQAv2} and 10\% \textbf{VizWiz} data. For LYP, the VQA model is trained on A+B and selector on A+B with annotations from 10 models.
}
\label{tab:vizwiz}
\end{table}

\begin{table}[t]
\centering
\footnotesize
\setlength{\tabcolsep}{2.3pt}
\begin{tabular}{lccccccc}
&&& \multicolumn{5}{c}{50\% OOD} \\
\cmidrule(l){4-8} 
Model &  $f$ train & $g$ train & AUC & \covatr{1} & \covatr{5} &  \covatr{10} & \grey{Acc.}  \\
\midrule
MaxProb & A & -- & 25.53 & 0.00 & 14.55 & 23.49 & \grey{48.07} \\
Selector & A & B & 22.83 & 9.31 & 21.87 & 32.02 & \grey{48.07}  \\
Selector + LYP & AB & AB & \textbf{22.71} & \textbf{11.81} & \textbf{22.51}  & \textbf{32.18} & \grey{48.14} \\
\end{tabular}
\caption{OFA-Base results with 50\% \textbf{VQAv2} and 50\% \textbf{VizWiz} data. For LYP, the VQA model is trained on A+B and selector on A+B with annotations from 10 models.
}
\label{tab:vizwiz3}
\end{table}

\section{Selective Prediciton for Visual Entailment}\label{sec:snlive}

We show experiments for our models and LYP on the visual entailment dataset SNLI-VE~\cite{xie2019snlive} in \cref{tab:snlive}. Given an image premise, and a text hypothesis, the model has to return one of the three possible outputs: \textit{entailment}, \textit{neutral}, or \textit{contradiction}. We run the experiments on OFA-Base, and use the same setup as the original OFA paper \cite{wang2022ofa}, except that we do not use the textual premise to make it comparable to previous works. 
We divide the SNLI-VE training set into 80\% for the A split, and the remaining 20\% for the B split. We use the original validation and test splits for model selection and test.

We see that LYP is very effective on this task: it improves the coverage across all risk levels compared to MaxProb and Selector baselines.

\begin{table}[t]
\centering
\setlength{\tabcolsep}{2.8pt}
\footnotesize
\begin{tabular}{clcccccc}
$f$ train & Model & $g$ train & AUC & \covatr{1} & \covatr{5} & \covatr{10} &  \grey{Acc} \\
\midrule
\multirow{2}{*}{A} & MaxProb & -- & 10.10 & 6.69 & 28.08 & 49.75 & \grey{77.31} \\
& Selector & B & 8.86 & 13.19 & 38.51  & 59.29 & \grey{77.24} \\ \midrule
\multirow{3}{*}{AB} & MaxProb  & -- & 8.78 & 12.04 & 35.44 & 59.56  & \grey{77.88} \\  & Selector & AB (Self) & 8.49 & 12.62  & 37.69 & 61.64  & \grey{77.91} \\  & Selector & AB (LYP) & \textbf{8.22} & \textbf{16.57} & \textbf{40.02} & \textbf{62.90}  & \grey{77.91} \\ 
\end{tabular}
\caption{OFA-Base results on SNLI-VE (without textual premise).}
\label{tab:snlive}
\end{table}

\section{Qualitative Examples}
\label{sec:app:qualitative}

\begin{figure*}[t]
    \centering
    \includegraphics[width=0.92\linewidth]{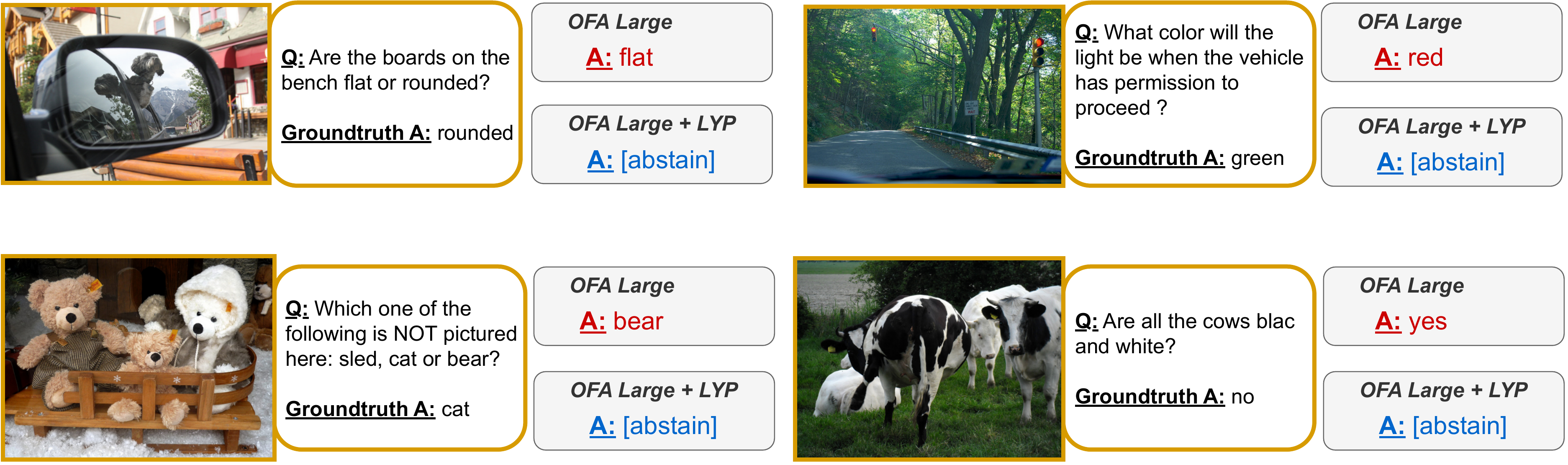}
    \caption{Qualitative examples for \ofalarge on \advqa: On those examples, the baseline (\maxprob) answers incorrectly the answer, and our model with \ours abstains. For both models, the threshold is selected on in-distribution data for maximum coverage at 5\% risk.}         \label{fig:advqa:abstain}
\end{figure*}

\begin{figure*}[t]
    \centering
    \includegraphics[width=0.92\linewidth]{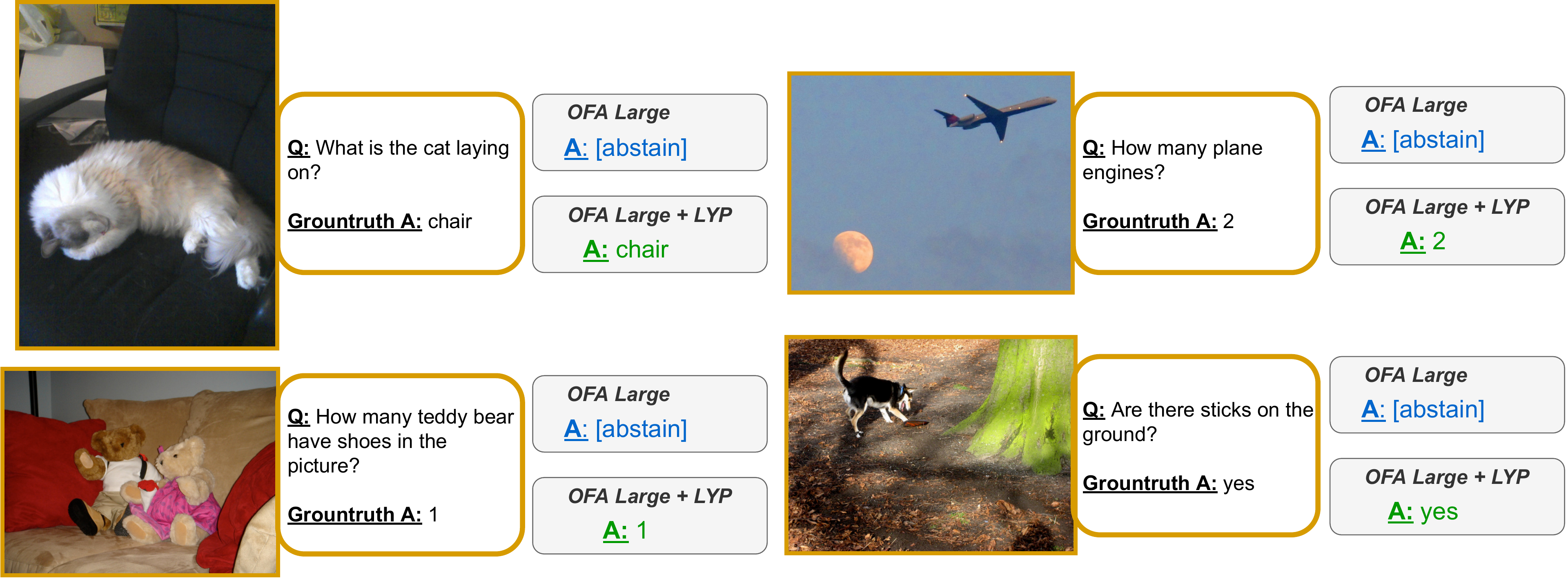}
    \caption{Qualitative examples for \ofalarge on \advqa: On those examples, the baseline model abstains but had predicted the correct answer. \ofalarge + \ours does not abstain. The threshold is selected on in-distribution data for maximum coverage at 5\% risk.}
    \label{fig:advqa:correct}
\end{figure*}

\begin{figure*}[t]
    \centering
    \includegraphics[width=0.92\linewidth]{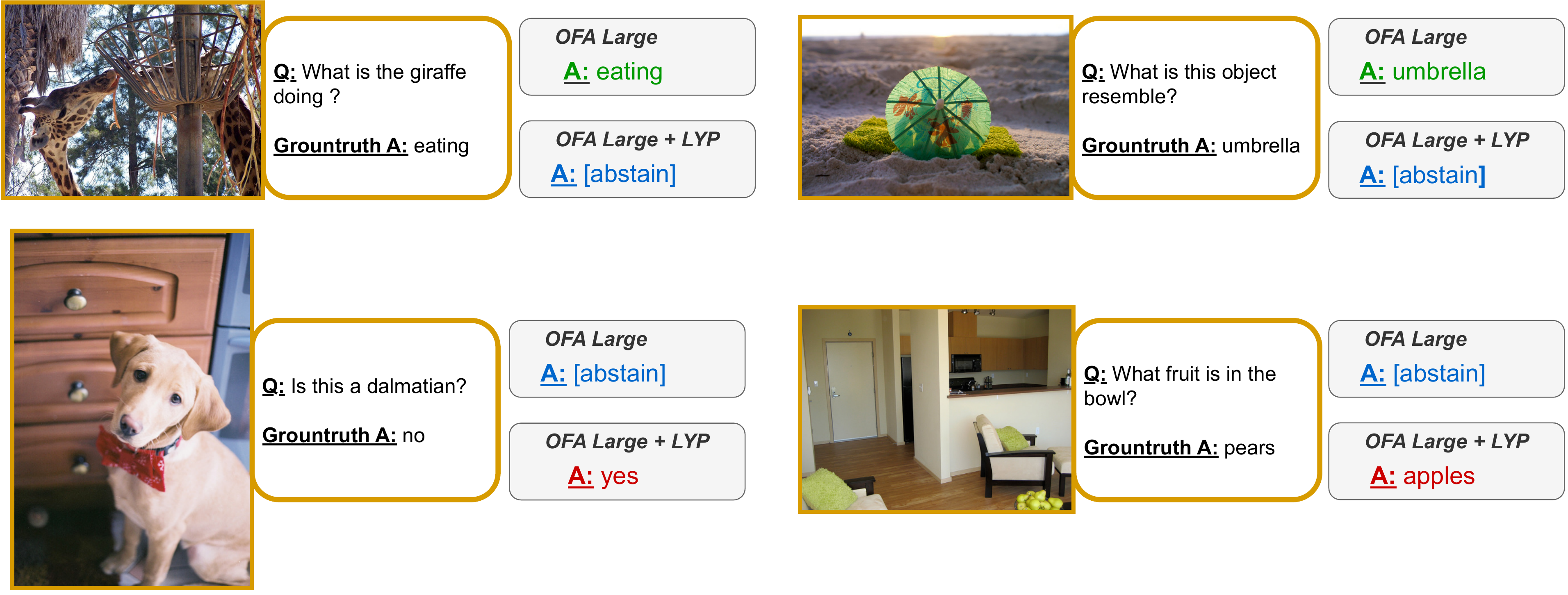}
    \caption{Failure cases for \ofalarge + \ours on \advqa: On the first two examples, the baseline predicts the correct answer, and \ofalarge + \ours abstains. On the second line, the baseline abstains from answering an incorrect answer, while \ofalarge + \ours still answers.
    For both models, the threshold is selected on in-distribution data for maximum coverage at 5\% risk.}
    \label{fig:advqa:incorrect}
\end{figure*}

\cref{fig:advqa:abstain,fig:advqa:correct,fig:advqa:incorrect} show qualitative results comparing the \ofalarge + \ours and \ofalarge + MaxProb, on the AdVQA dataset.
In both cases, the \ofalarge model $f$ is trained on \traindev.
For all examples, the abstention threshold is set on the in-distribution validation set to get maximum coverage at 5\% risk.

\cref{fig:advqa:abstain} shows examples where the VQA model (OFA-Large) is incorrect.
Thus, the correct behavior is to abstain. But the MaxProb model does not abstain using the provided threshold, instead, it answers incorrectly. On the contrary, our model OFA-L + \ours abstains.

\cref{fig:advqa:correct} shows examples where the OFA-L model is correct: the best behavior is to answer. The MaxProb model abstains, while our method answers correctly.

\cref{fig:advqa:incorrect} shows two kinds of failure cases of our models: In the first line, OFA-L + \ours incorrectly abstains, as the VQA model was correct. In the second line, our model incorrectly answer instead of abstaining, as the answer provided by the model was incorrect.

\renewcommand{\midsplitcline}{\arrayrulecolor{lightgray}\cmidrule(r){2-2}\cmidrule(lr){3-5} \cmidrule(lr){6-6}  \cmidrule(lr){7-9} \cmidrule(lr){10-10} \cmidrule(lr){11-13}\arrayrulecolor{black}}

\begin{table*}[t]
\centering
\footnotesize
\begin{tabular}{lcl@{}clrrrrrrrr}
\toprule
\multicolumn{2}{c}{VQA Model $f$} & \multicolumn{3}{c}{Selection func. $g$}   & \multirow{2}{*}{ \grey{Acc $\uparrow$}} & \multicolumn{3}{c}{\symcovatr in $\% \uparrow$} & \multirow{2}{*}{AUC $\downarrow$} & \multirow{2}{*}{\effrel{1}} & \multirow{2}{*}{\effrel{10}} & \multirow{2}{*}{\effrel{100}}  \\
\cmidrule(lr){1-2} \cmidrule(lr){3-5} \cmidrule(lr){7-9}
Name & Train Set &  Name & Train Set & Targets& & \covatr{1} & \covatr{5} & \covatr{10} \\
\midrule
\multirow{5}{*}{CLIP-ViL} & \multirow{2}{*}{\train}
& MaxProb & - & - \hspace{10pt}\cite{whitehead2022reliable}  & \grey{69.98} & 4.97 &33.79 &53.62 &10.92 &54.67 &21.40 &1.32 \\
&& Selector & \dev & Self \cite{whitehead2022reliable} & \grey{69.98} & 15.79 & 37.79 & 55.65 & 10.21 & 55.44 & 25.82 & 8.74 \\
\midsplitcline
& \multirow{3}{*}{\traindev} 
& MaxProb  & - & - & \grey{70.72} & 5.54 & 34.84 & 55.04 & 10.49 & 55.93 & 22.81 & 2.59 \\
&& Selector & \traindev & Self & \grey{70.72} & 6.45 & 34.26 & 56.07 & 10.48 & 56.07 & 22.99 & 2.39 \\
&& Selector & \traindev & \ours & \grey{70.72}  & \textbf{18.40} & \textbf{38.65} & \textbf{57.40} & \textbf{9.76} & \textbf{56.53} & \textbf{26.45} & \textbf{9.74} \\ 
\midrule
\multirow{5}{*}{\ofabase} & \multirow{2}{*}{\train} 
& MaxProb & - & - & \grey{74.87} & 3.45 & 45.60 & 66.61 & 7.99 & 62.52 & 30.57 & 6.81 \\ && Selector & \dev & Self & \grey{74.87} & 23.78 & 49.16 & 69.00 & 7.32 & 63.03 & 34.39 & 12.53 \\
&& Selector & \traindev & LYP-A Self-B & \grey{74.87} & 26.03 & 51.18 & 69.97 & 7.13 & 63.48 & 35.91 & 14.38\\
\midsplitcline
& \multirow{3}{*}{\traindev}
& MaxProb & - & - & \grey{75.18} & 14.88 & 46.15 & 67.51 & 7.79 & 63.04 & 30.13 & 7.29\\ && Selector & \traindev &  Self & \grey{75.18} & 26.64 & 50.80 & 69.56 & 7.10 & 63.66 & 34.92 & 12.92 \\ 
&& Selector & \traindev & \ours & \grey{75.18} & \textbf{27.71} & \textbf{51.64} & \textbf{70.20} & \textbf{6.98} & \textbf{63.88} & \textbf{36.29} & \textbf{16.30}\\
\midrule
\multirow{5}{*}{\ofalarge } & \multirow{2}{*}{\train} 
& MaxProb  & - & - & \grey{77.53} & 20.57 & 53.99 & 75.18 & 6.42 & 66.68 & 36.12 & 8.21 \\
&& Selector & \dev & Self & \grey{77.53} & 30.86 & 58.05 & 76.65 & 5.81 & 67.34 & 41.43 & 17.58 \\
&& Selector & \traindev & LYP-A Self-B & \grey{77.53} & 32.05 & 59.05 & 77.10 & 5.69 & 67.61 & 42.10 & 18.55 \\
\midsplitcline
& \multirow{3}{*}{\traindev} 
& MaxProb & - & - &  \grey{77.79} & 16.31 & 53.83 & 75.27 & 6.43 & 66.96 & 36.06 & 6.29 \\
&& Selector & \traindev &  Self &  \grey{77.79} & 31.47 & 58.80 & 77.14 & 5.69 & 67.82 & 41.43 & 16.08 \\
&& Selector & \traindev & \ours & \grey{77.79} & \textbf{32.92} & \textbf{59.43} & \textbf{77.52} & \textbf{5.60} & \textbf{68.02} & \textbf{42.83} & \textbf{18.78} \\
\bottomrule
\end{tabular}
\caption{
Risk-coverage metrics and effective reliability on ID data (i.e., \vqatwo test split~\cite{whitehead2022reliable}). Scores for \ofalarge are averaged over 5 trials. This table is a copy from the main paper with the additional lines ``LYP-A Self-B'', discussed in \cref{sec:supp:LYPASelfB}.}
\label{tab:supp:main-results}
\end{table*}

\begin{table*}[t!]
\centering
\footnotesize
\begin{tabular}{lcl@{}clrrrrrrrr}
\toprule
\multicolumn{2}{c}{VQA Model $f$} & \multicolumn{3}{c}{Selection function $g$}    & \multirow{2}{*}{ \grey{Acc $\uparrow$}} & \multicolumn{3}{c}{$\mathcal{C} @ \mathcal{R}$ in $\% \uparrow$} & \multirow{2}{*}{ AUC $\downarrow$} & \multirow{2}{*}{\effrel{1}} & \multirow{2}{*}{\effrel{10}} & \multirow{2}{*}{\effrel{100}} \\
\cmidrule(lr){1-2} \cmidrule(lr){3-5} \cmidrule(lr){7-9}
Name & Train Set&  Name & Train Set & Targets& & \covatr{1} & \covatr{5} & \covatr{10} \\
\midrule  
\multirow{5}{*}{ CLIP-ViL } 
    & \multirow{2}{*}{\train}
& MaxProb & - & - \hspace{10pt}\cite{whitehead2022reliable} & \grey{66.35} & 0.00 & 24.16 & 43.53 & 13.55 & 49.12 & 14.39 & -4.64 \\
&& Selector & \dev & Self~\cite{whitehead2022reliable} &    \grey{66.35} & 12.69 & 31.12 & 46.96 & 12.47 & 50.36 & 20.15 & 5.22 \\
\midsplitcline
& \multirow{3}{*}{\traindev} 
& MaxProb  & - & - & \grey{67.12} & 2.60 & 26.13 & 45.25 & 12.97 & 50.49 & 16.59 & -0.93 \\
&& Selector & \traindev &  Self & \grey{67.12} & 2.97 & 26.70 & 46.19 & 12.80 & 50.89 & 18.19 & -0.65 \\
&& Selector & \traindev & \ours & \grey{67.12} & \textbf{15.22} & \textbf{32.58} & \textbf{49.18} & \textbf{11.90} & \textbf{51.43} & \textbf{22.09} & \textbf{7.12} \\
\midrule
\multirow{5}{*}{\ofabase } & \multirow{2}{*}{\train} 
& MaxProb & - & - &  \grey{71.59} & 0.01 & 36.07 & 56.49 & 10.10 & 57.49 & 23.15 & -0.34 \\
&& Selector & \dev & Self & \grey{71.59} & 18.32 & 41.48 & 59.74 & 9.19 & 57.97 & 27.22 & 9.09 \\
&& Selector & \traindev & LYP-A Self-B & \grey{71.61} & 19.49 & 43.04 & 61.04 & 9.00 & 58.43 & 29.23 & 7.68 \\
\midsplitcline
& \multirow{3}{*}{\traindev} 
& MaxProb & - & - & \grey{72.00} & 1.74 & 37.02 & 57.57 & 9.78 & 58.11 & 22.09 & 0.53 \\
&& Selector & \traindev &  Self & \grey{72.00} & 19.72 & 42.70 & 60.84 & 8.90 & 58.90 & 28.05 & 2.88 \\
&& Selector & \traindev & \ours & \grey{72.00} & \textbf{21.58} & \textbf{44.09} & \textbf{61.69} & \textbf{8.74} & \textbf{59.11} & \textbf{28.79} & \textbf{10.88} \\
\midrule
\multirow{5}{*}{\ofalarge } & \multirow{2}{*}{\train} 
& MaxProb & - & - & \grey{74.56} & 4.76 & 44.53 & 66.06 & 8.21 & 61.90 & 28.20 & 0.21 \\
&& Selector & \dev & Self & \grey{74.56} & 23.53 & 50.17 & 68.76 & 7.33 & 62.96 & 34.43 & 9.88 \\
&& Selector & \traindev & LYP-A Self-B  & \grey{74.56} & 24.34 & 50.78 & 69.46 & 7.25 & 63.15 & 34.79 & \textbf{11.55} \\
\midsplitcline
& \multirow{3}{*}{\traindev} 
& MaxProb & - & - & \grey{74.79} & 1.30 & 43.70 & 65.95 & 8.26 & 62.24 & 27.09 & -2.46  \\
&& Selector & \traindev & Self & \grey{74.79} & 22.68 & 50.27 & 69.27 & 7.32 & 63.03 & 33.50 & 4.92 \\
&& Selector & \traindev & \ours & \grey{74.79} & \textbf{25.38} & \textbf{51.07} & \textbf{69.74} & \textbf{7.17} & \textbf{63.41} & \textbf{34.85} & 10.34 \\
\bottomrule
\end{tabular}
\caption{Mixed ID/OOD scenario, composed of 90\% \vqatwo and 10\% \advqa examples. 
This table is a copy from the main paper with the additional lines ``LYP-A Self-B'', discussed in \cref{sec:supp:LYPASelfB}.
}
\label{tab:supp:main-advqa}
\end{table*}

\begin{table*}[t]
\centering
\footnotesize
\begin{tabular}{lcl@{}clrrrrrrrr}
\toprule
\multicolumn{2}{c}{VQA Model $f$} & \multicolumn{3}{c}{Selection function $g$}    &  \multirow{2}{*}{ \grey{Acc $\uparrow$}} & \multicolumn{3}{c}{\symcovatr in $\% \uparrow$} & \multirow{2}{*}{ AUC $\downarrow$} & \multirow{2}{*}{\effrel{1}} & \multirow{2}{*}{\effrel{10}} &  \multirow{2}{*}{\effrel{100}} \\
\cmidrule(lr){1-2} \cmidrule(lr){3-5} \cmidrule(lr){7-9}
Name & Train Set&  Name & Train Set & Targets& & \covatr{1} & \covatr{5} & \covatr{10}\\
\cmidrule(r){1-2}\cmidrule(lr){3-5} \cmidrule(lr){6-6}  \cmidrule(lr){7-9} \cmidrule(lr){10-10} \cmidrule(lr){11-13}
\multirow{6}{*}{ CLIP-ViL } & \multirow{2}{*}{\train}
& MaxProb & - & - \hspace{10pt}\cite{whitehead2022reliable}& \grey{58.36} & 0.00 & 7.08 & 21.97 & 20.62 & 36.59 & -1.47 & -14.38 \\
&& Selector & \dev & Self \cite{whitehead2022reliable} & \grey{58.36} & 5.87 & 17.41 & 29.21 & 18.90 & 38.76 & 7.11 & -2.20 \\
\arrayrulecolor{lightgray} \midsplitcline \arrayrulecolor{black}
& \multirow{3}{*}{\traindev} 
& MaxProb  & - & - & \grey{59.29} & 1.11 & 10.17 & 24.99 & 19.58 & 38.42 & 2.99 & -9.79\\
&& Selector & \traindev &  Self & \grey{59.29} & 0.07 & 11.21 & 25.86 & 19.28 & 39.17 & 5.90 & -7.37 \\
&& Selector & \traindev & \ours & {\grey{59.29}} & \textbf{7.07} & \textbf{19.13} & \textbf{31.53} & \textbf{17.94} & \textbf{39.85} & \textbf{12.67} & \textbf{3.40} \\
\midrule
\multirow{6}{*}{\ofabase } & \multirow{3}{*}{\train}
& MaxProb & - & - & \grey{64.17} & 0.01 & 18.83 & 34.15 & 15.71 & 46.05 & 5.33 & -28.66  \\
&& Selector & \dev & Self  & \grey{64.17} & 6.97 & 25.19 & 39.53 & 14.44 & 46.41 & \textbf{11.98} & -3.47 \\
&& Selector & \traindev & LYP-A Self-B & \grey{64.16} & 6.76 & 25.39 & 41.49 & 14.26 & 46.99 & 14.42 & -9.89 \\
\midsplitcline
& \multirow{3}{*}{\traindev} 
& MaxProb & - & - & \grey{64.63} & 0.03 & 17.57 & 33.94 & 15.43 & 46.32 & 2.11 & -19.21 \\
&& Selector & \traindev &  Self & \grey{64.63} & 5.11 & 25.83 & 40.13 & 14.09 & 47.58 & 10.75 & -21.18 \\
&& Selector & \traindev & \ours  & \grey{64.63} & \textbf{9.41} & \textbf{27.89} & \textbf{42.00} & \textbf{13.80} & \textbf{48.03} & 11.89 & \textbf{-2.81}\\
\midrule
\multirow{6}{*}{\ofalarge } & \multirow{3}{*}{\train} 
& MaxProb & - & - & \grey{67.78} & 0.39 & 22.67 & 43.82 & 13.05 & 50.88 & 10.05 & -20.68 \\
&& Selector & \dev & Self & \grey{67.79} & 9.18 & 32.42 & 50.06 & \textbf{11.60} & 52.74 & 18.31 & -12.07 \\
&& Selector & \traindev & LYP-A Self-B & \grey{67.79} & \textbf{11.24} & \textbf{31.01} & \textbf{50.37} & 11.65 & \textbf{52.76} & \textbf{18.47} & \textbf{-6.53} \\
\midsplitcline
& \multirow{3}{*}{\traindev} 
& \maxprob & - & - & \grey{67.78} & 0.13 & 21.01 & 42.31 & 13.37 & 51.02 & 7.58 & -26.60 \\
&& Selector & \traindev &  Self & \grey{67.77} & 6.58 & 30.15 & 48.83 & 11.98 & 51.79 & 15.37 & -23.92 \\
&& Selector & \traindev & \ours & \grey{67.77} & 9.44 & 30.87 & 49.69 & 11.75 & 52.51 & 16.95 & -12.91 \\
\bottomrule
\end{tabular}
\caption{Results on a mixed ID/OOD setting, composed of 66.7\% \vqa data (Test split in \cref{tab:dataset-size}) and 33.3\% AdVQA examples. Discussion in \cref{sec:appendix:OODresults}.}
\label{tab:main-advqa-33}
\end{table*}

\begin{table*}[t]
\centering
\footnotesize
\begin{tabular}{lcl@{}clrrrrrrrr}
\toprule
\multicolumn{2}{c}{VQA Model $f$} & \multicolumn{3}{c}{Selection function $g$}    &  \multirow{2}{*}{ \grey{Acc $\uparrow$}} & \multicolumn{3}{c}{\symcovatr in $\% \uparrow$} & \multirow{2}{*}{ AUC $\downarrow$} & \multirow{2}{*}{\effrel{1}} & \multirow{2}{*}{\effrel{10}} &  \multirow{2}{*}{\effrel{100}} \\
\cmidrule(lr){1-2} \cmidrule(lr){3-5} \cmidrule(lr){7-9}
Name & Train Set&  Name & Train Set & Targets& & \covatr{1} & \covatr{5} & \covatr{10} \\
\cmidrule(r){1-2}\cmidrule(lr){3-5} \cmidrule(lr){6-6}  \cmidrule(lr){7-9} \cmidrule(lr){10-10} \cmidrule(lr){11-13}
\multirow{6}{*}{ CLIP-ViL } & \multirow{2}{*}{\train}
& MaxProb & - & - \hspace{10pt}\cite{whitehead2022reliable}& \grey{52.66} & 0.00 & 3.08 & 9.77 & 26.57 & 27.65 & -13.20 & -20.60\\
&& Selector & \dev & Self \cite{whitehead2022reliable}& \grey{52.66} & \textbf{4.19} & 10.29 & 18.17 & 24.49 & 30.62 & -2.24 & -7.94\\
\midsplitcline
& \multirow{3}{*}{\traindev} 
& MaxProb  & - & - & \grey{53.83} & 0.97 & 3.66 & 12.27 & 25.23 & 29.82 & -6.40 & -15.61 \\
&& Selector & \traindev &  Self & \grey{53.83} & 0.04 & 5.52 & 13.38 & 24.96 & 30.82 & -2.96 & -11.50\\
&& Selector & \traindev & \ours & \grey{53.83} & 3.41 & \textbf{11.19} & \textbf{20.42} & \textbf{23.22} & \textbf{31.85} & \textbf{5.49} & \textbf{-0.04} \\
\midrule
\multirow{6}{*}{\ofabase } & \multirow{3}{*}{\train} 
& MaxProb & - & - & \grey{59.17} & 0.01 & 5.78 & 18.48 & 20.80 & 38.14 & -6.11 & -29.98 \\
&& Selector & \dev & Self  & \grey{59.18} & 3.21 & 15.98 & 26.27 & 19.00 & 38.49 & 0.10 & -11.07 \\
&& Selector & \traindev & LYP-A Self-B & \grey{59.18} & 2.28 & 15.38 & 26.72 & {18.80} & 39.24 & \textbf{3.46} & -19.92 \\
\midsplitcline
& \multirow{3}{*}{\traindev} 
& MaxProb & - & - & \grey{59.61} & 0.06 & 6.91 & 20.86 & 20.17 & 38.45 & -12.19 & -31.48\\
&& Selector & \traindev &  Self & \grey{59.62} & 2.29 & 15.78 & 27.38 & 18.70 & 39.88 & -0.74 & -36.10 \\
&& Selector & \traindev & \ours  & \grey{59.62} & \textbf{3.98} & \textbf{17.13} & \textbf{28.53} & \textbf{18.30} & \textbf{40.35} & -0.49 & \textbf{-11.27} \\
\midrule
\multirow{6}{*}{\ofalarge } & \multirow{3}{*}{\train} 
& MaxProb & - & - & \grey{63.02} & 0.31 & 11.53 & 27.85 & 17.18 & 43.42 & -3.11 & -34.01 \\
&& Selector & \dev & Self & \grey{63.01} & \textbf{5.56} & \textbf{20.11} & \textbf{35.51} & \textbf{15.52} & 45.49 & \textbf{6.48} & -26.57 \\
&& Selector & \traindev & LYP-A Self-B & \grey{63.01} & 5.07 & 19.65 & 34.80 & 15.61 & \textbf{45.55} & 6.11 & \textbf{-18.23} \\
\midsplitcline
& \multirow{3}{*}{\traindev} 
& \maxprob & - & - & \grey{62.93} & 0.12 & 6.22 & 26.58 & 17.58 & 43.40 & -6.02 & -40.93 \\
&& Selector & \traindev &  Self & \grey{62.93} & 1.00 & 18.55 & 33.48 & 16.03 & 44.14 & 2.57 & -43.03 \\
&& Selector & \traindev & \ours & \grey{62.93} & 3.51 & 19.74 & 34.18 & 15.78 & 45.03 & 4.19 & -29.46 \\
\bottomrule
\end{tabular}
\caption{Results on a mixed ID/OOD setting, composed of 50\% \vqa data (Test split in \cref{tab:dataset-size}) and 50\% AdVQA examples. Discussion in \cref{sec:appendix:OODresults}.
}
\label{tab:main-advqa-50}
\end{table*}

\begin{table*}[t]
\centering
\footnotesize
\begin{tabular}{lcl@{}clrrrrrrrr}
\toprule
\multicolumn{2}{c}{VQA Model $f$} & \multicolumn{3}{c}{Selection function $g$}    &  \multirow{2}{*}{ \grey{Acc $\uparrow$}} & \multicolumn{3}{c}{\symcovatr in $\% \uparrow$} & \multirow{2}{*}{ AUC $\downarrow$} & \multirow{2}{*}{\effrel{1}} & \multirow{2}{*}{\effrel{10}} &  \multirow{2}{*}{\effrel{100}} \\
\cmidrule(lr){1-2} \cmidrule(lr){3-5} \cmidrule(lr){7-9}
Name & Train Set&  Name & Train Set & Targets& & \covatr{1} & \covatr{5} & \covatr{10} \\
\cmidrule(r){1-2}\cmidrule(lr){3-5} \cmidrule(lr){6-6}  \cmidrule(lr){7-9} \cmidrule(lr){10-10} \cmidrule(lr){11-13}
\multirow{6}{*}{ CLIP-ViL } & \multirow{2}{*}{\train}
& MaxProb & - & - \hspace{10pt}\cite{whitehead2022reliable} & \grey{46.66} & 0.00 & 0.00 & 3.04 & 33.67 & 18.32 & -24.68 & -28.56\\
&& Selector & \dev & Self \cite{whitehead2022reliable} & \grey{46.66} & 1.91 & 5.65 & 10.09 & 31.43 & 22.00 & -11.50 & -12.05\\
\midsplitcline
& \multirow{3}{*}{\traindev} 
& MaxProb  & - & - & \grey{47.94} & 0.67 & 1.28 & 5.59 & 32.08 & 20.87 & -16.85 & -21.99\\
&& Selector & \traindev &  Self & \grey{47.94} & 0.05 & 1.44 & 5.49 & 31.79 & 22.20 & -11.69 & -15.28\\
&& Selector & \traindev & \ours & \grey{47.94} & \textbf{2.13} & \textbf{6.60} & \textbf{10.44} & \textbf{29.77} & \textbf{23.60} & \textbf{-0.77} & \textbf{-0.89} \\
\midrule
\multirow{6}{*}{\ofabase } & \multirow{2}{*}{\train} 
& MaxProb & - & - & \grey{53.71} & 0.00 & 0.45 & 8.44 & 26.47 & 29.64 & -17.60 & -43.15  \\
&& Selector & \dev & Self & \grey{53.77} & 2.00 & 8.23 & 15.97 & 24.45 & 30.21 & -10.46 & -16.01 \\
&& Selector & \traindev & LYP-A Self B & \grey{53.77} & 1.73 & 7.79 & 15.51 & 24.35 & 30.86 & \textbf{-7.20} & -28.39 \\
\midsplitcline
& \multirow{3}{*}{\traindev} 
& MaxProb & - & - &  \grey{54.28} & 0.03 & 0.53 & 10.16 & 25.72 & 29.96 & -25.56 & -44.75 \\
&& Selector & \traindev &  Self & \grey{54.26} & 1.52 & 8.79 & 16.23 & 24.15 & 31.88 & -12.68 & -52.56 \\
&& Selector & \traindev & \ours  & \grey{54.26} & \textbf{1.95} & \textbf{9.71} & \textbf{17.11} & \textbf{23.79} & \textbf{32.38} & -12.12 & \textbf{-20.65} \\
\midrule
\multirow{6}{*}{\ofalarge } & \multirow{2}{*}{\train} 
& MaxProb & - & - & \grey{57.69} & 0.13 & 3.65 & 14.24 & 22.36 & 34.91 & -16.36 & -49.70 \\
&& Selector & \dev & Self & \grey{57.71} & \textbf{3.03} & \textbf{11.20} & \textbf{22.04} & \textbf{20.44} & 37.45 & \textbf{-5.27} & -39.01 \\
&& Selector & \traindev & LYP-A Self-B & \grey{57.71} & 1.89 & 10.78 & 20.09 & 20.63 & \textbf{37.51} & -6.12 & \textbf{-30.14} \\
\midsplitcline
& \multirow{3}{*}{\traindev} 
& MaxProb & - & - & \grey{57.52} & 0.08 & 0.54 & 13.41 & 22.87 & 34.70 & -20.37 & -56.21 \\
&& Selector & \traindev &  Self &  \grey{57.50} & 0.46 & 9.02 & 20.14 & 21.10 & 35.39 & -10.72 & -61.53 \\
&& Selector & \traindev & \ours & \grey{57.50} & 0.08 & 10.28 & 19.93 & 20.94 & 36.60 & -8.58 & -44.52 \\
\bottomrule
\end{tabular}
\caption{Results on a mixed ID/OOD setting, composed of 33.3\% \vqa  data  (Test split in \cref{tab:dataset-size}) and 66.7\% AdVQA examples. Discussion in \cref{sec:appendix:OODresults}.
}
\label{tab:main-advqa-66}
\end{table*}

\renewcommand{\midsplitcline}{\arrayrulecolor{lightgray}
\cmidrule(r){2-2}\cmidrule(lr){3-5} \cmidrule(lr){6-6}  \cmidrule(lr){7-8} \cmidrule(lr){9-10} \cmidrule(lr){11-12}\arrayrulecolor{black}}

\begin{table*}[h!]
\small
\centering
\begin{tabular}{lcl@{}clrrrrrrr}
\toprule
\multicolumn{2}{c}{VQA Model $f$} & \multicolumn{3}{c}{Selection function $g$}     & \multirow{2}{*}{\grey{Acc $\uparrow$}} & \multicolumn{2}{c}{$\mathcal{R}=1\%$}  & \multicolumn{2}{c}{$\mathcal{R}=5\%$}  & \multicolumn{2}{c}{$\mathcal{R}=10\%$} \\
\cmidrule(lr){1-2} \cmidrule(lr){3-5} \cmidrule(lr){7-8} \cmidrule(lr){9-10} \cmidrule(lr){11-12}
Name & Train Set&  Name & Train Set & Targets & &  \multicolumn{1}{c}{ $\mathcal{R}$} & \multicolumn{1}{c}{ $\mathcal{C}$} & \multicolumn{1}{c}{ $\mathcal{R}$} &  \multicolumn{1}{c}{ $\mathcal{C}$} & \multicolumn{1}{c}{ $\mathcal{R}$} &  \multicolumn{1}{c}{ $\mathcal{C}$} \\
\cmidrule(r){1-2}\cmidrule(lr){3-5} \cmidrule(lr){6-6}  \cmidrule(lr){7-8} \cmidrule(lr){9-10} \cmidrule(lr){11-12}
\multirow{5}{*}{CLIP-ViL} & \multirow{2}{*}{\train} 
& MaxProb & - & - \hspace{12pt}\cite{whitehead2022reliable} & \grey{69.98} & 0.86 & 3.49 & 4.55 & 31.59 & 9.60 & 52.35 \\ & & Selector & \dev &  Self \cite{whitehead2022reliable} & \grey{69.98} & 0.72 & 13.26 & 4.74 & 36.66 & 9.97 & 55.58 \\ \midsplitcline
& \multirow{3}{*}{\traindev}
& MaxProb & - & - & \grey{70.72} & 1.08 & 6.67 & 4.59 & 32.85 & 9.83 & 54.47 \\ & & Selector & \traindev & Self & \grey{70.72} & 1.10 & 7.60 & 4.78 & 34.16 & 9.73 & 54.63 \\ & & Selector & \traindev & LYP &  \grey{70.72} & 0.85 & 16.78 & 4.96 & 38.30 & 10.08 & 57.34 \\ \midrule
\multirow{6}{*}{\ofabase} & \multirow{2}{*}{\train} 
& MaxProb & - & - & \grey{74.87} & 1.18 & 5.32 & 4.96 & 45.45 & 9.96 & 66.44 \\ & & Selector & \dev &  Self &  \grey{74.87} & 1.05 & 24.54 & 5.07 & 49.53 & 10.18 & 69.67 \\
\midsplitcline
& \multirow{3}{*}{\traindev} 
& MaxProb & - & - & \grey{75.18} & 0.82 & 4.32 & 4.98 & 46.03 & 10.08 & 67.88 \\ && Selector & \traindev &  Self & \grey{75.18} & 1.14 & 27.88 & 5.23 & 51.76 & 10.09 & 69.87 \\ && Selector & \traindev & \ours  & \grey{75.18} & 1.00 & 27.84 & 5.17 & 52.44 & 10.35 & 71.31 \\ \midrule
\multirow{6}{*}{\ofalarge} & \multirow{2}{*}{\train} 
& MaxProb & - & - & \grey{77.53} & 0.99 & 20.46 & 4.95 & 53.73 & 9.80 & 74.51 \\
& & Selector & \dev &  Self & \grey{77.53} & 1.10 & 32.01 & 5.04 & 58.23 & 9.98 & 76.63 \\
\midsplitcline
& \multirow{3}{*}{\traindev} 
& MaxProb & - & - & \grey{77.80} & 1.01 & 15.06 & 4.85 & 53.11 & 9.83 & 74.67 \\
&& Selector & \traindev &  Self & \grey{77.79} & 1.00 & 31.45 & 4.94 & 58.57 & 9.97 & 77.08 \\
&& Selector & \traindev & \ours  & \grey{77.79} & 0.99 & 32.79 & 4.99 & 59.39 & 10.05 & 77.67 \\
\bottomrule
\end{tabular}
\caption{Results on the ID \vqa evaluation set (Test split in \cref{tab:dataset-size}). Thresholds for desired risk level are selected on the in-distribution \val split. Discussion in \cref{sec:thresholdgeneralization}.}
\label{tab:generalization:ID}
\end{table*}

\begin{table*}[h!]
\small
\centering
\begin{tabular}{lcl@{}clrrrrrrr}
\toprule
\multicolumn{2}{c}{VQA Model $f$} & \multicolumn{3}{c}{Selection function $g$} & \multirow{2}{*}{\grey{Acc $\uparrow$}} & \multicolumn{2}{c}{$\mathcal{R}=1\%$}  & \multicolumn{2}{c}{$\mathcal{R}=5\%$}  & \multicolumn{2}{c}{$\mathcal{R}=10\%$} \\
\cmidrule(lr){1-2} \cmidrule(lr){3-5} \cmidrule(lr){7-8} \cmidrule(lr){9-10} \cmidrule(lr){11-12}
Name & Train Set&  Name & Train Set & Targets & &  \multicolumn{1}{c}{ $\mathcal{R}$} & \multicolumn{1}{c}{ $\mathcal{C}$} & \multicolumn{1}{c}{ $\mathcal{R}$} &  \multicolumn{1}{c}{ $\mathcal{C}$} & \multicolumn{1}{c}{ $\mathcal{R}$} &  \multicolumn{1}{c}{ $\mathcal{C}$} \\
\cmidrule(r){1-2}\cmidrule(lr){3-5} \cmidrule(lr){6-6}  \cmidrule(lr){7-8} \cmidrule(lr){9-10} \cmidrule(lr){11-12}
\multirow{5}{*}{CLIP-ViL} & \multirow{2}{*}{\train} 
& MaxProb & - & - \hspace{12pt}\cite{whitehead2022reliable} & \grey{66.35} & 1.83 & 3.21 & 6.25 & 29.53 & 12.05 & 50.06 \\
& & Selector & \dev &  Self \cite{whitehead2022reliable} & \grey{66.35} & 0.95 & 12.14 & 5.75 & 33.92 & 11.78 & 52.33 \\
\midsplitcline
& \multirow{3}{*}{\traindev}
& MaxProb & - & - & \grey{67.12} & 1.59 & 6.11 & 5.97 & 30.70 & 12.02 & 52.14 \\
& & Selector & \traindev & Self & \grey{67.12} & 1.52 & 6.97 & 6.04 & 31.95 & 11.63 & 51.83 \\
& & Selector & \traindev & LYP & \grey{67.12} & 1.14 & 15.26 & 5.81 & 35.46 & 11.72 & 54.08 \\
\midrule
\multirow{5}{*}{\ofabase} & \multirow{2}{*}{\train} 
& MaxProb & - & - & \grey{71.59} & 1.69 & 4.88 & 6.54 & 43.00 & 12.11 & 64.13 \\
& & Selector & \dev &  Self & \grey{71.60} & 1.43 & 22.60 & 6.19 & 46.18 & 12.23 & 67.04 \\
\midsplitcline
& \multirow{3}{*}{\traindev} 
& MaxProb & - & - & \grey{72.00} & 1.30 & 3.95 & 6.56 & 43.59 & 12.05 & 65.67\\
&& Selector & \traindev &  Self & \grey{72.02} & 1.60 & 25.72 & 6.49 & 48.75 & 11.82 & 67.13 \\
&& Selector & \traindev & \ours & \grey{72.01} & 1.38 & 25.61 & 6.27 & 48.97 & 12.07 & 68.25 \\
\midrule
\multirow{5}{*}{\ofalarge} & \multirow{2}{*}{\train} 
& MaxProb & - & - & \grey{74.56} & 1.58 & 18.93 & 6.50 & 51.43 & 11.96 & 73.01 \\
& & Selector & \dev &  Self & \grey{74.56} & 1.56 & 29.66 & 6.23 & 55.37 & 11.84 & 74.44 \\
\midsplitcline
& \multirow{3}{*}{\traindev} 
& MaxProb & - & - &  \grey{74.79} & 1.57 & 13.90 & 6.50 & 50.93 & 12.00 & 73.16 \\
&& Selector & \traindev &  Self & \grey{74.79} & 1.52 & 29.05 & 6.34 & 55.91 & 11.90 & 75.17 \\
&& Selector & \traindev & \ours  & \grey{74.79} & 1.47 & 30.17 & 6.29 & 56.31 & 11.99 & 75.66 \\
\bottomrule
\end{tabular}
\caption{Results on the mixed 90\% \vqa + 10\% AdVQA evaluation set (\vqa data is from the Test split in \cref{tab:dataset-size}).  Thresholds for desired risk level are selected on our in-distribution \val set. Discussion in \cref{sec:thresholdgeneralization}.}
\label{tab:generalization:90}
\end{table*}

\end{document}